\documentclass[pdflatex,sn-basic]{sn-jnl}
\usepackage{graphicx}%
\usepackage{multirow}%
\usepackage{amsmath,amssymb,amsfonts}%
\usepackage{amsthm}%
\usepackage{mathrsfs}%
\usepackage[title]{appendix}%
\usepackage{xcolor}%
\usepackage{textcomp}%
\usepackage{manyfoot}%
\usepackage{booktabs}%
\usepackage{algorithm}%
\usepackage{algorithmicx}%
\usepackage{algpseudocode}%
\usepackage{listings}%
\usepackage{microtype}      
\usepackage{xcolor}         
\usepackage{multirow}
\usepackage{textcomp}%
\usepackage{manyfoot}%
\usepackage{booktabs}%
\usepackage{listings}%
\usepackage{threeparttable}

\usepackage{makecell}
\usepackage{array}

\usepackage{bbding}
\usepackage{rotating}
\usepackage{pifont}
\usepackage{hyperref}
\usepackage{changes}
\usepackage{colortbl}
\usepackage{changes}

\usepackage{paralist}
\usepackage[edges]{forest}
\usepackage{siunitx}
\usepackage{tikz-dependency}
\usepackage{tikz}
\usepackage{tcolorbox} 
\usepackage{arydshln}

\begin{document}

\title[From Pixel to Coding: Evaluating the Figure Reproduction Capabilities of MLLMs]{From Pixel to Coding: Evaluating the Figure Reproduction Capabilities of MLLMs}

\author[1,2]{\fnm{Zijian} \sur{Chen}}\email{zijian.chen@sjtu.edu.cn}

\author[1]{\fnm{Zhengyu} \sur{Chen}}\email{chenzy828@sjtu.edu.cn}

\author[2]{\fnm{Bohan} \sur{Liang}}\email{bobliang@sjtu.edu.cn}

\author[3]{\fnm{Lirong} \sur{Deng}}\email{p2523406@mpu.edu.mo}

\author[1,2]{\fnm{Yushuo} \sur{Zheng}}\email{yushuozheng@sjtu.edu.cn}

\author[1,2]{\fnm{Yanwei} \sur{Jiang}}\email{jiang-yan-wei@sjtu.edu.cn}

\author[2]{\fnm{Qi} \sur{Jia}}\email{jiaqi@pjlab.org.cn}

\author[2]{\fnm{Kaiwei} \sur{Zhang}}\email{zhangkaiwei@pjlab.org.cn}

\author[1]{\fnm{Wenjun} \sur{Zhang}}\email{zhangwenjun@sjtu.edu.cn}

\author[1,2,*]{\fnm{Guangtao} \sur{Zhai}}\email{zhaiguangtao@sjtu.edu.cn}

\affil[1]{\orgdiv{Institute of Image Communication and Information Processing}, \orgname{Shanghai Jiao Tong University}, \orgaddress{\city{Shanghai}, \postcode{200240}, \country{China}}}

\affil[2]{\orgname{Shanghai Artificial Intelligence Laboratory}, \orgaddress{\city{Shanghai}, \postcode{200030}, \country{China}}}

\affil[3]{\orgname{Macao Polytechnic University}, \orgaddress{\city{Macao}, \postcode{999078}, \country{China}}}

\affil[*]{\orgname{Corresponding author: Guangtao Zhai}}

\abstract{
Multimodal Large Language Models (MLLMs) have demonstrated impressive capabilities in both visual understanding and code generation. However, existing benchmarks typically evaluate these two modalities in isolation, lacking a dedicated assessment of their unification, i.e., how a model can perceive complex visual structures and synthesize them into precise, executable code. Moreover, current visual code generation benchmarks often rely on simplified layouts within single programming environments, falling short of evaluating true unified multimodal reasoning.
To bridge this gap, we propose FigCodeBench, a comprehensive framework for rigorously evaluating MLLMs on figure reproduction, integrating multimodal comprehension and generation. 
We first design a systematic dataset construction pipeline, resulting in a total of 6,194 instances that cover 7 functional categories and 4 types of programming languages (Python, Matlab, R, and Latex). 
We further categorize figure reproduction into three tiers (Easy, Medium, Hard) with visual and code complexity modeling, specifically targeting complex structural reasoning, varying aspect ratios, and dense geometric constraints. 
We introduce a multi-dimensional evaluation protocol, encompassing visual fidelity (e.g., PSNR, SSIM, LPIPS) and syntactic isomorphism (e.g., abstract syntax tree similarity), that aligns highly with the Mean Machine Opinion Score (MMOS) and human preferences. 
Based on our framework, we conducted extensive experiments on 24 widely used proprietary and open-source MLLMs (e.g., Gemini 3.1 Pro, GPT-5.4, and Kimi-K2.5), where we observed a universal, non-linear performance cliff across different programming languages and difficulty scenarios for all models, and gained several insights, such as the significant metric decline in rigid declarative languages. 
Our benchmark offers (1) a comprehensive dataset and multi-dimensional evaluation pipeline; (2) an up-to-date leaderboard on MLLM figure reproduction proficiency; and (3) a nuanced understanding of the perceiving and reasoning bottlenecks hindering unified comprehension and generation in current MLLMs.
}


\maketitle

\section{Introduction}
Multimodal Large Language Models (MLLMs) have demonstrated extraordinary advances in image understanding and code generation. Current models \citep{xu2025qwen3,qwen35,zeng2026glm,grok43,gpt54} can now effortlessly complete Visual Question Answering (VQA) \citep{liu2024mmbench} and achieve near-perfect accuracy on visual math problems \citep{wang2024measuring}, coding challenges \citep{chen2021evaluating,jimenez2024swe}, and strong performance on more recent multimodal reasoning benchmarks such as RealWorldQA \citep{realworldqa} and OmniDocBench 1.5 \citep{ouyang2025omnidocbench}.
Recent releases touted as vision-language breakthroughs, such as Gemini 3.1 Pro \citep{gemini31pro} and Qwen3.8-Max \citep{qwen38}, have pushed performance towards the upper limits of these benchmarks. 
For example, Qwen3.8-Max achieves over 95\% accuracy on MathVision \citep{wang2024measuring} and over 92\% pass@1 on LiveCodeBench v6 \citep{jainlivecodebench}. 

Yet, existing evaluations predominantly operate within isolated paradigms that evaluate modalities in a disjointed manner, failing to assess the true unification of comprehension and generation. On one hand, traditional code generation benchmarks \citep{jainlivecodebench, zhenglivecodebench} rely exclusively on textual specifications as input, neglecting the reality that real-world programming is inherently a multimodal endeavor, where developers frequently reference visual materials. 
On the other hand, existing multimodal evaluations \citep{yue2025mmmu,fu2026ocrbench,zhang2025large} are largely confined to perceptual comprehension, which is typically assessed through VQA or multiple-choice formats. They fall short of validating a model's capacity for further structural representation, leaving it unknown whether an MLLM can accurately synthesize its visual understanding into complex, executable generative outputs. 
Benchmarks like CharXiv \citep{wang2024charxiv} offer visual chart comprehension challenges, but suffer from passive question-answering, rely on rigid text-matching metrics susceptible to superficial guessing, lack executable validation, and the inability to verify active figure reproduction.

To bridge this gap, some pioneering efforts have begun to explore cross-modal visual code generation, tasking models with translating images into executable scripts. 
Plot2Code \citep{wu2025plot2code}, ChartCoder \citep{zhao2025chartcoder}, and ChartMimic \citep{yangchartmimic} utilize information-intensive visual charts and textual instructions as inputs, requiring MLLMs to generate the corresponding code for chart rendering. Design2Code \citep{si2025design2code} assesses how well MLLMs can generate code that renders into webpages matching given reference screenshots.
BioMotion Arena \citep{chen2025can} draws inspiration from the remarkable capacity of human vision to perceive biological motions and prompts MLLMs to generate code for human animation that satisfies the combination of action requirements.
However, these benchmarks predominantly focus on overly simplified, single-language data plots or front-end web development (e.g., HTML/CSS rendering). 
They inherently lack the mathematical rigor, dense geometric constraints, and multi-paradigm complexity required for professional scientific figure reproduction. Furthermore, their evaluation protocols often rely on unidimensional visual similarity or heuristic script matching, failing to capture the true structural fidelity across diverse programming environments.

\begin{figure*}
\centerline{\includegraphics[width=1\textwidth]{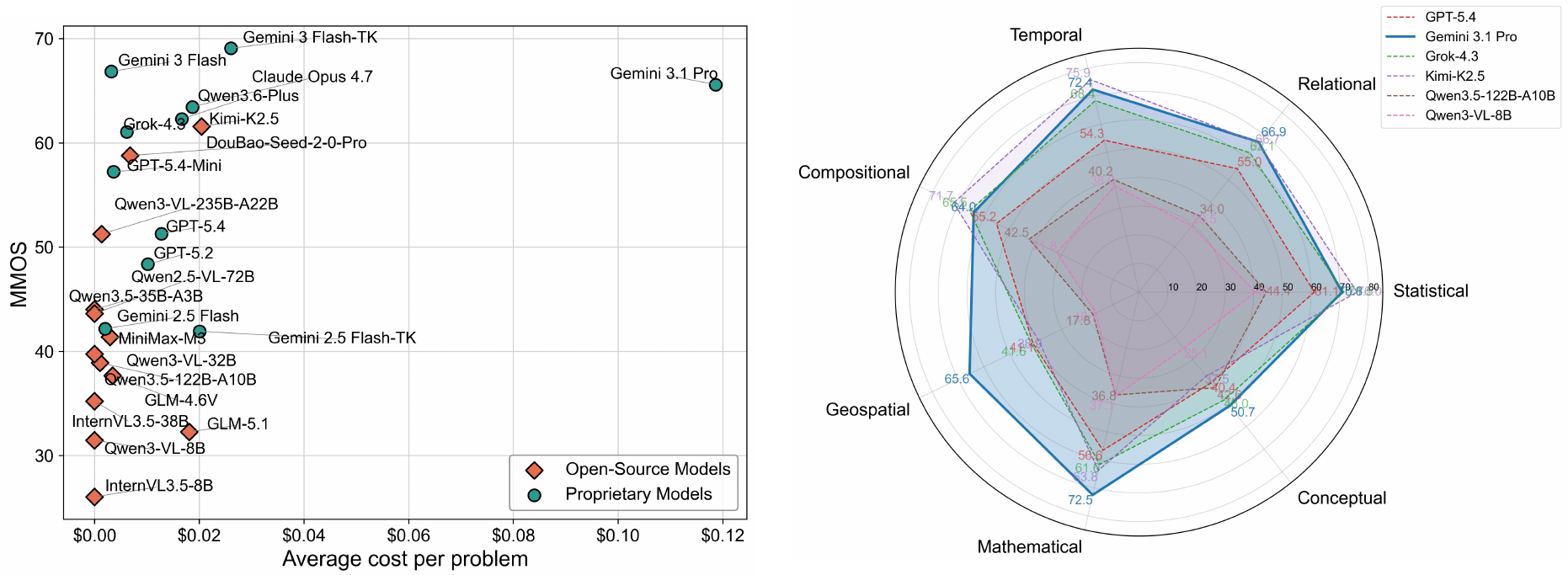}}
\caption{{\bf FigCodeBench leaderboard}. {\it Left}: Mean Machine Opinion Score (MMOS) versus average cost per problem for various models. {\it Right}: Performance comparison of six representative models on different image categories.}
\label{leaderboard}
\end{figure*}

To this end, we introduce FigCodeBench, a visual programming benchmark for figure reproduction, which challenges MLLMs to generate the corresponding code for reproducing a given figure. FigCodeBench is characterized by its (1) real-world fidelity, (2) more programming languages, (3) diverse figure types, and (4) multi-dimensional evaluation metrics. 
Specifically, prior benchmarks predominantly focus on simplistic charts (e.g., standard bar or pie graphs), while FigCodeBench expands the horizon to complex scientific figures in different coding scenarios. Our benchmark comprehensively evaluates the model's ability to synthesize intricate typographic layouts, geospatial projections, mathematical vector fields, flowcharts, and other abstract diagrams, moving beyond simple chart reading to holistic code-driven scientific visualization.
Through the collection of academic documents and online repositories, we identify four common programming languages, i.e., Python, Matlab, R, and Latex, as well as 7 function-focused figure subcategories.
Subsequently, we manually collect a total of 6,194 high-quality figure-code pairs for these types with human-in-the-loop augmentation to mitigate data contamination. 
Furthermore, we establish a suite of evaluation metrics from both image-oriented and code-oriented perspectives to thoroughly assess the performance of MLLMs.
We also propose to extract the visual quality achieved relative to the amount of code required as a joint fidelity-efficiency metric that quantifies the reproduction efficiency.

We evaluate a suite of frontier models on FigCodeBench, including proprietary models such as Gemini 3.1 Pro, GPT-5.4, Claude Opus 4.7, and open-source models such as Kimi-K2.5 and Qwen3.5-122B (Fig. \ref{leaderboard}). 
We observe a significant performance gap across different programming languages, where rendering via descriptive Latex constitutes a formidable bottleneck for all evaluated models.
In-depth experiments on different image types further reveal that current MLLMs lack the 2D spatial reasoning capabilities required for code-based object depiction and the representation of abstract set relationships, once detached from a regular grid coordinate system.
Meanwhile, we find a significant divergence in the coding strategies adopted by different MLLMs, highlighting that few models can strike a perfect balance between precise visual comprehension and reliable code generation.
We also compare reasoning models with their non-reasoning counterparts and identify the impact of model size in handling such problems.
Correlation analysis demonstrates a high correlation between our adopted metrics and human evaluation, validating the effectiveness of these metrics.
Further case study reveals that although current MLLMs have largely mastered basic data-to-geometry mapping, they still exhibit substantial weaknesses in fine-grained aesthetic rendering.

Overall, our contributions can be summarized as follows:
\begin{itemize}
    \item This paper introduces FigCodeBench, a comprehensive framework designed for evaluating the unified visual comprehension and code generation capabilities of MLLMs, consisting of a large-scale figure reproduction dataset and a multi-perspective evaluation protocol.
    \item We design a systematic dataset construction pipeline resulting in 4 programming languages and 7 functional categories with 6,194 multimodal figure-code pairs, alongside a multi-dimensional metric system that provides reliable assessments highly aligned with both machine judgments and human evaluations.
    \item We conduct large-scale experiments on 11 proprietary and 13 open-source MLLMs, revealing that existing models still face substantial challenges in figure reproduction tasks. We observe widespread performance cliffs across scenarios and derive several insights into the perceptual and generative bottlenecks of current MLLMs.
\end{itemize}

\section{Related Work}\label{sec::rw}
\subsection{Multimodal Large Language Models}
Multimodal Large Language Models (MLLMs) \citep{hurst2024gpt,li2024llava,chen2024internvl,team2023gemini,team2024gemini,wang2024qwen2,gemini25pro} have demonstrated remarkable performance across a broad spectrum of tasks, evolving from foundational visual perception to complicated multimodal reasoning across images, video, and 3D space \citep{chen2025just,fei2024video,wuspatial}. 
The integration of the visual capabilities of vision transformers (e.g., QwenViT \citep{bai2025qwen3}) and the generative capabilities of LLMs enables the seamless bridging of raw visual perception with high-level cognitive reasoning, allowing models to interpret complex visual scenes and follow open-ended instructions. 
Notably, in the AI for Science (AI4S) domain, the Intern-S1 series \citep{bai2025intern,zou2026intern} equips an MoE LLM with a vision encoder, a time-series encoder, and a dynamic tokenizer that switches the tokenization and embedding strategies for natural language and scientific inputs, achieving leading results across scientific domains (e.g., chemistry, materials, life-science, earth, etc.).
In the medicine domain, Med-MLLM \citep{liu2023medical} supports medical data across visual modalities (e.g., chest X-ray and CT) and textual modalities (e.g., medical reports and free-text clinical notes), thereby facilitating downstream clinical tasks. 
For more routine user interface (GUI) automation control, CoCo-Agent \citep{ma2024coco} and MP-GUI \citep{wang2025mp} enhance GUI perception across different aspects and granularities, including screenshots and detailed layouts for the visual channel, and historical actions for the textual channel.
Recently, building upon the rapid advancements in the understanding and reasoning capabilities of unimodal large models, researchers have further extended MLLMs to audio \citep{cappellazzo2025large,Qwen3-TTS}, code \citep{cao2026qwen3,codex}, action \citep{qi2025llavaction,szot2024grounding}, and even Omni modalities \citep{xu2025qwen3,jiang2025specific}, which further expanded the application scope of MLLMs.

\subsection{Evaluation of MLLMs}
Early multimodal benchmarks, such as VQA v2 \citep{goyal2017making}, LVLM-eHub \citep{xu2024lvlm}, MME \citep{fu2025mmecomprehensiveevaluationbenchmark}, and Seed-Bench \citep{li2024seed}, primarily focus on Visual Question Answering (VQA). Later, a greater number of visual-language tasks have emerged to assess MLLMs’ capabilities across various dimensions, including Optical Character Recognition (OCR) \citep{liu2023hidden}, chart understanding \citep{masry2022chartqa,wang2024charxiv}, abstract visual reasoning \citep{chen2025can,zhang2025puzzlebench}, long video reasoning \citep{wu2024longvideobench,wang2025lvbench}, scientific and multidisciplinary tasks \citep{zhouscientists,yue2025mmmu,chenobi,chen2025pictobi,rong2025liveproteinbench,zheng2026benchmarking}. 
In contrast, our FigCodeBench bridges the evaluation of the perception capabilities of MLLMs with the code generation abilities of their language components, enabling a cross-modality evaluation that aligns with the practical capabilities required in actual scientific analysis workflows.

\subsection{Code Generation Benchmarks} 
Natural language to code generation has gained growing attention recently, with a proliferation of code-specific models \citep{fakhoury2024llm,cao2026qwen3,gpt53codex,cursor}, necessitating the development of more comprehensive and diverse evaluation benchmarks. 
For example, HumanEval \citep{chen2021evaluating}, HumanEval-XL \citep{peng2024humaneval}, and DSCodeBench \citep{ouyang2026dscodebench} focus on code generation for various programming languages.
LiveBench \citep{whitelivebench} assesses a model’s ability to parse a coding competition question statement and write a correct answer, with completion ability concerned.
LiveCodeBench \citep{jainlivecodebench} expands the usage scenarios of code LLMs to code generation, self-repair, code execution, and test case output prediction. 
LiveCodeBench Pro \citep{zhenglivecodebench} further elevates code problems to human grandmaster levels, featuring sophisticated algorithmic reasoning.
SWE-Bench \citep{jimenezswe}, on the other hand, focuses more on assessing software engineering scenarios for code maintenance rather than algorithm design.
However, these benchmarks predominantly serve LLMs, leaving MLLMs insufficiently scrutinized regarding the intricate synergy between visual grounding and code generation.

\begin{table*}[!t]
  \caption{{\bf FigCodeBench statistics} for annotated tags with corresponding functional-focus taxonomy. Breakdown by the number of figure plotting topics, query image counts, percentages, average code length ({\it tokens calculated by the LLama3 tokenizer}), and difficulty scores. The slash separates statistics of exemplary data ({\it left}) and user-generated data ({\it right}).}
  \label{statistics}
  \centering
  \resizebox{1\linewidth}{!}{\begin{tabular}{lllcrrrr}
    \toprule
    {\bf Language} &{\bf Package}    & {\bf Tag}  & {\bf \# Prototype}  & {\bf \# Query Imgs}&{\bf Pct.\%}&{\bf Code Len.}&{\bf Difficulty} \\
    \midrule
    \multirow{6}{*}{Python} &\multirow{6}{*}{\makecell[l]{Matplotlib\\\&Seaborn}}&Statistical &56/51& 168/148&9.72/3.31 & 379.7/772.4&[30.52,60.25]/[2.42,61.69]  \\
    &&Relational &29/44 &87/131&5.03/2.93&  337.3/625.2&[31.35,60.92]/[36.72,64.56] \\
    &&Temporal &17/34 &51/97&2.95/2.17 &  377.1/514.2&[38.09,52.42]/[16.76,59.65] \\
    &&Compositional &9/8 &27/24 &1.56/0.54& 305.9/485.1& [39.13,52.50]/[43.46,52.31]\\
    &&Geospatial &8/1 &24/3 &1.39/0.07& 338.6/566.3 &[34.39,55.02]/[44.87,50.88] \\
    &&Mathematical &37/39 &111/116 &6.42/2.60&376.2/496.1&[32.40,55.47]/[41.84,60.91]  \\
    \midrule 
    \multirow{6}{*}{Matlab}&\multirow{6}{*}{Built-in}&Statistical&8/33 &24/98 &1.39/2.19&508.9/704.5&[33.33,54.25]/[36.73,57.21]   \\
    &&Relational &8/21 &21/77 &1.21/1.72&482.8/663.0 &[32.38,52.61]/[42.48,56.92]  \\
    &&Temporal &12/25 & 35/75&2.02/1.68&451.2/842.8 &[34.48,51.12]/[38.98,58.41]  \\
    &&Compositional &5/11&13/33 &0.75/0.74 & 388.2/1005.4& [38.80,50.14]/[46.62,61.78] \\
    &&Geospatial &3/4 &9/10&0.52/0.22 &615.4/193.1& [41.14,54.01]/[33.18,47.44]  \\
    &&Mathematical &35/34 &104/101&6.02/2.26 & 475.7/789.8 &[31.94,55.38]/[43.57,59.08] \\
    \midrule
    \multirow{6}{*}{R}&\multirow{6}{*}{ggplot2}&Statistical&47/97&141/289 &8.16/6.47& 310.3/717.9 &[28.08,54.43]/[34.18,63.13]   \\
    &&Relational &33/41 &99/122&5.73/2.73 &341.0/1108.2&[31.66,53.65]/[36.83,61.15]   \\
    &&Temporal &26/43 &78/126&4.51/2.82 &407.1/1032.1&[34.03,52.34]/[40.84,72.86]   \\
    &&Compositional &13/37 &39/107 &2.26/2.40&297.8/768.2&[37.77,52.39]/[38.61,61.25]   \\
    &&Geospatial &4/0 &12/0 &0.69/0.00&181.2/0.0&[18.68,42.35]/[0.00,0.00] \\
    &&Mathematical &15/24 & 45/72&2.60/1.61&398.0/613.9&[34.56,49.27]/[38.39,59.52]   \\
    \midrule
    Latex &TikZ &  Conceptual  &285/1,345&641/2,836 &37.07/63.52 &1344.2/1668.6&[27.83,76.98]/[25.33,73.99]\\
    \midrule
    {\bf Total}&&&651/1,897&1,729/4,465&100\%/100\%&\textendash&$[2.42,76.98]$\\
    \bottomrule
  \end{tabular}}
\end{table*}

\section{Benchmark Curation}
\label{sec::Curation}
In this section, we first introduce the design philosophy of FigCodeBench (Sec. \ref{subsec::philosophy}), and elaborate on the data curation process (Sec. \ref{subsec::data}). Then, we quantify the data diversity and difficulty tiers (Sec. \ref{subsec::division}). Afterwards, we define the evaluation taxonomy (Sec. \ref{subsec::taxonomy}) and make comparisons with existing related benchmarks (Sec. \ref{subsec::comparisons}).

\subsection{Design Philosophy}
\label{subsec::philosophy}
FigCodeBench is designed to evaluate MLLMs' ability to generate the precise programmatic representation of a scientific visualization from its visual manifestation, thereby differentiating their visual perception, coding, and cross-modal reasoning skills. 
Currently, the lack of systematic, reproducible benchmarks hinders progress in scientific visualization models.
Specifically, an input tuple $\mathcal{X} = \{I, L, A\}$ includes the target reference image $I$ to be reproduced, the target programming language $L$, and a set of auxiliary instructions $A$, providing textual context or specific constraints (e.g., {\it dataset} or {\it package requirements}). Given $\mathcal{X}$, the model $M_\theta$ aims to generate code $C$ by estimating the conditional probability distribution:
\begin{equation}
    C \sim P_{M_\theta}(C \mid I, L, A)
\end{equation}
Afterwards, the corresponding code compiler maps the generated script $C$ to a rendered image $\hat{I}$, which should be semantically and visually consistent with $I$ across multiple fidelity dimensions.
Guided by this objective, we construct the benchmark around two core principles.

{\bf Comprehensiveness and Fidelity.}
Unlike existing chart-related MLLM benchmarks \citep{yang2024matplotagent,wu2025plot2code,yangchartmimic} that aim at Python only, the tasks in FigCodeBench further cover three mainstream programming languages for figure visualization, i.e., Matlab, R, and LaTeX. 
These languages dominate the landscape of scientific computing and academic publishing, yet they exhibit fundamentally different syntactic structures, rendering paradigms, and API logic. By expanding the evaluation scope across these diverse compiler spaces, FigCodeBench rigorously assesses the unified multimodal comprehension and generation abilities of MLLMs, mitigating the risk of evaluating models that merely overfit to the idiosyncratic APIs of a single ecosystem. 
Furthermore, to guarantee real-world fidelity, the reference figures and user queries in our benchmark are strictly curated from authentic, in-the-wild sources rather than being synthetically generated via predefined templates.


{\bf Reproducibility and Determinism.}
Reproducibility is a core design criterion. Each task is anchored by an explicit visual outcome that serves as the basis for evaluation.
The execution workflows for all four programming spaces are encapsulated within a containerized sandbox environment. By strictly fixing random seeds and version-controlling the backend rendering engines and their auxiliary libraries, we ensure a one-to-one mapping between a given script and its generated figure and avoid stochastic model behavior. Meanwhile, FigCodeBench is designed with a robust, multidimensional evaluation framework that equips objective measurements and a multimodal LLM jury mechanism to ensure highly reproducible, fair, and definitive assessments. 
Beyond visual and semantic accuracy, we incorporate understanding-generation efficiency as a fundamental facet of our evaluation protocol, which is assessed through execution cost, including money, token expenses, and consistency measures, to quantify variability and facilitate repeated and reproducible evaluations across the research community.

\begin{figure*}
\centerline{\includegraphics[width=1\textwidth]{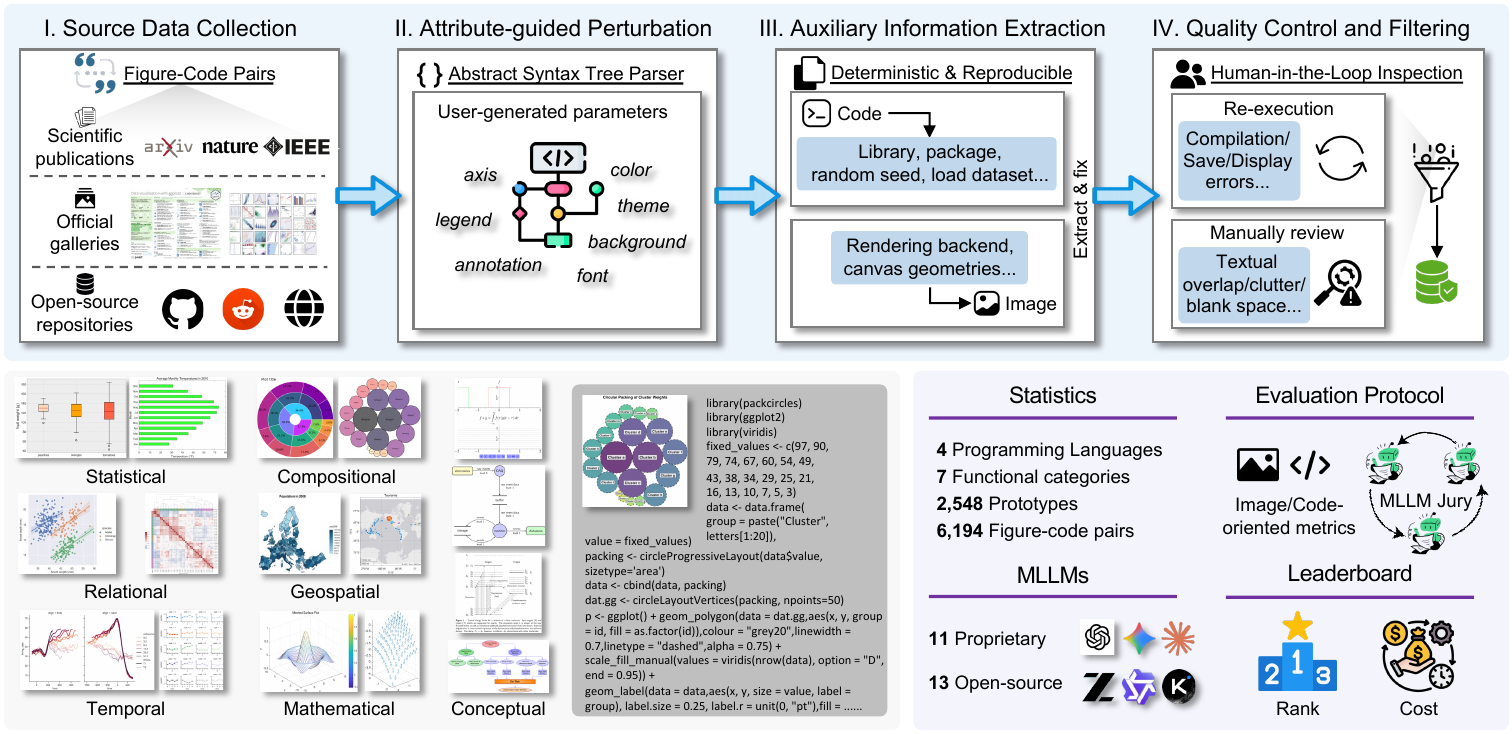}}
\caption{{\bf Overview of FigCodeBench}. {\it Upper}: The construction pipeline of FigCodeBench includes source data collection, attribute-guided perturbation, auxiliary information extraction, and quality control. {\it Bottom left}: This gallery displays 16 representative cases, organized into seven functional categories, i.e., statistical, relational, temporal, compositional, geospatial, mathematical, and conceptual visualizations. We summarize the key features of this benchmark in the {\it bottom right} part.}
\label{pipeline}
\end{figure*}

\subsection{Data Collection Pipeline}
\label{subsec::data}
As illustrated in Fig. \ref{pipeline}, constructing FigCodeBench progresses through four distinct stages: figure-code scraping, diversity perturbation, auxiliary information extraction, and quality control. Here, we provide an overview.

{\bf Data Sources.} We collect source figures and their generating codes from arXiv, peer-reviewed scientific publications, professional visualization galleries (e.g., Matplotlib, Seaborn, TikZ, PGFPlots, and ggplot2), and open-source repositories (e.g., GitHub and Reddit) that hold a CC BY 4.0 license, resulting in an initial collection of over 4,100 figures.
The collected instances exhibit the genuine complexities of professional scientific communication. This includes complex multi-axis layouts, specific visual representations, sophisticated aesthetic configurations (e.g., customized legends, non-standard aspect ratios, and perceptually uniform color schemes), and dense textual annotations.

{\bf Function-focused Taxonomy and Tags.}
Instead of categorizing figures solely by their morphological appearance, we identify 7 functional categories that span the common requirements of scientific communication, i.e., statistical (e.g., distributions and boxplots), relational (e.g., scatter and radar charts), temporal (e.g., time-series and area plots), Compositional (e.g., pie and tree maps), geospatial (e.g., map projections), mathematical (e.g., 3D surfaces and vector fields), and conceptual (e.g., flowcharts and various patterns).
Furthermore, each sample is annotated with the programming space (i.e., Python, LaTeX, R, and Matlab), the expertise level of the source code (i.e., exemplary and user-generated), and inherent complexity metrics, serving a critical diagnostic purpose to decouple baseline API memorization from authentic compositional generalization.  
We list the distribution of each tag in Tab. \ref{statistics}.

{\bf Attribute-guided Perturbation.} 
To mitigate the risk of data contamination, we augment the seed instances through controlled, attribute-guided perturbations. 
We develop an automated Abstract Syntax Tree (AST) parser and regex-based pipeline to isolate and modify specific semantic or stylistic attributes, such as axis, color, themes, legend, annotation, background, and fonts, within the source scripts, generating figure-code variants. 
We also invited five scientific researchers to manually contribute instances with personal preferences through an interactive graphical user interface (GUI) for human-guided curation. 
For each seed figure-code pair, we create 2 additional augmented pairs. 
Ultimately, this perturbation strategy enhances our dataset, yielding over 12,000 pairs that reflect a wide range of realistic and practical chart use cases.

{\bf Auxiliary Information Extraction.}
Beyond the source code, ensuring absolute reproducibility requires isolating the execution from environmental variance. To guarantee that the generation process is hardware-agnostic and strictly deterministic, we extract and explicitly anchor a comprehensive set of auxiliary information for each instance.
First, we identify and lock explicit dependencies, including third-party packages, library versions, and random seeds. 
More importantly, we resolve implicit rendering ambiguities by specifying standard graphic backends (e.g., applying a specific backend in Matplotlib to prevent interactive rendering collapse), enforcing uniform canvas geometries (e.g., DPI and aspect ratio configurations), and standardizing font assets to prevent layout corruption caused by Operating System (OS)-level font fallbacks.

{\bf Quality Control and Filtering.}
In this phase, we deploy a two-stage quality control and filtering process. First, all scripts are injected into isolated, predefined sandbox environments encompassing standardized compiler toolchains. Any instance that triggers compilation errors (e.g., image-saving errors caused by interactive maps), missing dependencies, or runtime exceptions is automatically pruned. 
Besides, we update the code blocks that use local data with simulated data to avoid errors stemming from unresolvable file I/O operations within the sandbox environment.
Second, we implement a human-in-the-loop inspection protocol. Domain experts systematically review the rendered images to identify and eliminate pathological visual artifacts that automated scripts might overlook, such as severe textual overlap, axis clutter, large areas of blank space, pure background, or disproportionate aspect ratios, guaranteeing that every reference figure in FigCodeBench represents a flawless, having scientific significance standard.
This process results in a refined collection of 6,194 figure-code pairs.

\begin{figure}[!t]
\centerline{\includegraphics[width=1\textwidth]{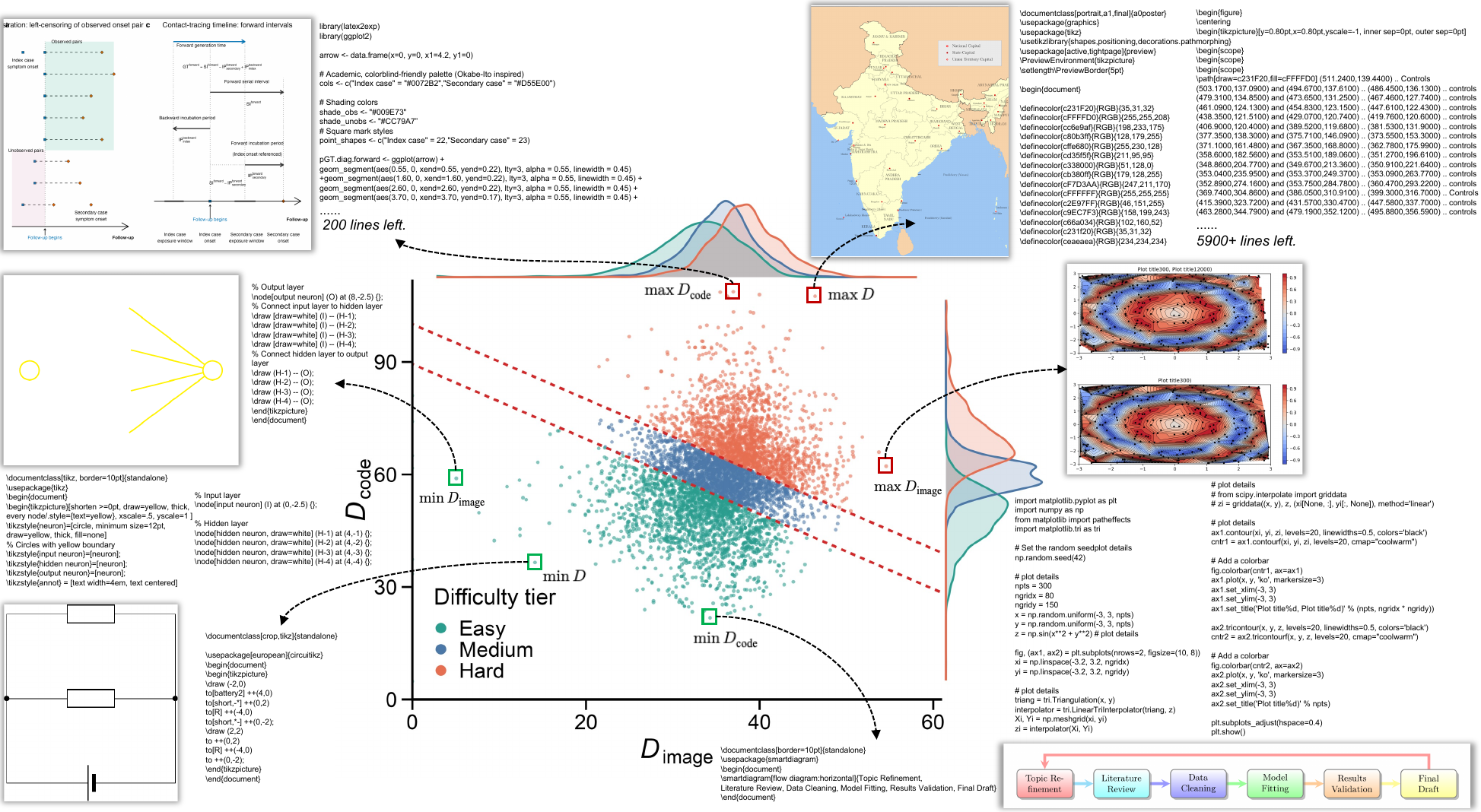}}
\caption{Scatter plot and marginal distribution of the image-code complexity of FigCodeBench. The red dotted lines separate the easy ($[2.42, 44.73]$), medium ($[44.73, 50.05]$) and hard tiers $[50.05, 76.98]$. We also visualize some image-code cases ({\it points that are neither zero-sum nor on the coordinate axes}) at the corresponding minimum and maximum difficulty tiers.}
\label{scatter_difficulty}
\end{figure}

\subsection{Data Statistics and Division}
\label{subsec::division}
As depicted in Tab. \ref{statistics}, FigCodeBench encompasses a total of 6,194 instances, covering 4 types of programming languages and 7 functional categories. Given the variety of their internal combination elements, we treat  2,548 unique instances as prototypes. 
We provide a detailed breakdown of dataset composition, including the number of query images, category proportions, and tokenized code lengths, for both exemplary and user-generated samples. Notably, the latter features longer code sequences and increased complexity, exhibiting the diversity of FigCodeBench.
Note that we do not deliberately impose sample uniformity, but rather reflect the usage frequency of different programming languages in specific scenarios.

{\bf Difficulty Tiers.}
To streamline navigation and ensure diversity, we design a figure-code complexity rating heuristic for difficulty tiers.
We quantify the intrinsic difficulty of each benchmark instance from two complementary perspectives, i.e., the visual complexity of the target figure and the structural complexity of its reference code. We characterize the visual complexity of the figure using five features, including edge density, line-orientation diversity, color diversity, number of foreground components, and foreground spatial occupancy:
\begin{itemize}
    \item {\bf Edge Density}: Edge density is defined as $F_{\mathrm{edge}}=N_{\mathrm{edge}}/{N_{\mathrm{valid}}}$, where $N_{\mathrm{edge}}$ is the number of Canny operator detected edge pixels and $N_{\mathrm{valid}}$ is the number of valid, non-padding pixels. A higher edge density generally indicates that the figure contains more curves, axes, text, markers, or other graphical structures. 
    \item {\bf Line-Orientation Diversity}: We compute horizontal and vertical image gradients using Sobel filters and evaluate their orientations only at Canny edge pixels. Gradient orientations are treated as unsigned line orientations and quantized into 18 equally spaced bins. Let $p_b$ denotes the proportion of edge pixels assigned to orientation bin $b$. The normalized orientation entropy is $F_{\mathrm{orientation}}=-\sum_{b=1}^{18} p_b \log p_b/{\log 18}$. The resulting value lies in $[0,1]$. A figure dominated by parallel or similarly oriented lines produces a low score, whereas a figure containing horizontal, vertical, curved, and oblique structures produces a high score. 
    \item {\bf Color Diversity}: Color diversity is measured using the entropy of a quantized CIELAB color histogram. The CIELAB space is divided into $8\times8\times8$ bins, giving $B=512$ possible color bins. Let $p_b$ denote the proportion of foreground pixels falling into color bin $b$. We define $F_{\mathrm{color}}=-\sum_{b=1}^{B} p_b \log p_b/{\log B}$. Consequently, a monochromatic line plot receives a lower score than a figure containing multiple colored curves, regions, or heatmaps \citep{rosenholtz2007measuring}.
    \item {\bf Number of Foreground Components}: We construct a foreground mask by comparing each valid pixel with the estimated background color $M(x,y)=\mathbf{1}\left[\Delta E\bigl(I(x,y),C_{\mathrm{bg}}\bigr)>\tau\right]$, where $\Delta E$ denotes the color difference in CIELAB space and the initial threshold is set to $\tau=5$. A $3\times3$ morphological closing operation is subsequently applied to connect small discontinuities, and connected regions containing fewer than 20 pixels are removed as noise. Let $N_{\mathrm{component}}$ denote the number of remaining foreground connected components. The component-complexity feature is defined as $F_{\mathrm{component}}=\log\left(1+N_{\mathrm{component}}\right)$, providing a representation of spatially separated image regions \citep{rosenfeld1966sequential}, where a larger number of components generally indicates more independent graphical elements, such as markers, text glyphs, annotations, and legend.
    \item {\bf Foreground Spatial Occupancy}: The spatial occupancy of a figure is defined as $F_{\mathrm{occupancy}}=N_{\mathrm{foreground}}/N_{\mathrm{valid}}$, where $N_{\mathrm{foreground}}$ is the number of foreground pixels in the cleaned mask and $N_{\mathrm{valid}}$ excludes padded pixels. This feature distinguishes visually sparse figures from figures whose graphical elements occupy a large fraction of the canvas.
\end{itemize}
To make the extracted visual features comparable across figures with different native resolutions, each ground-truth image is resized such that its longer side is 1,024 pixels while preserving the original aspect ratio. The resized image may be padded to a $1024\times1024$ canvas, but padded pixels are excluded from all subsequent measurements. For images containing transparency, the alpha channel is composited over a white background. We estimate the dominant background color, denoted by $C_{\mathrm{bg}}$, as the component-wise median of the CIELAB values of the pixels along the image boundary. Compared with using a fixed white background, this procedure is more robust to off-white canvases and other background colors.
The overall image-complexity score $D_{\mathrm{image}}$ is then defined as the averaged values of the five features and mapped to $[0,100]$. 

Before extracting code features, we remove comments, blank lines, import statements, package-loading commands, and automatically generated boilerplate. Large literal data blocks and long coordinate arrays are also excluded. These elements may substantially increase file length without increasing the complexity of the graphical construction procedure. The resulting representation is intended to quantify plotting logic rather than embedded data volume.
Here, we describe code complexity using three feature groups, including code size, Abstract Syntax Tree (AST) structure, and plotting-API usage.
\begin{itemize}
    \item {\bf Code Size}: We measure the number of effective lexical tokens as the code size $F_{\mathrm{size}}=\log\left(1+N_{\mathrm{eff.\ token}}\right)$. Effective tokens include function calls, operators, control-flow statements, parameters, and variable references.
    \item {\bf AST Structural Complexity}: The abstract syntax tree provides a language-aware representation of the syntactic organization of a program. We use both the total number of AST nodes and the maximum AST depth, i.e., $F_{\mathrm{AST}}=0.7\log\left(1+N_{\mathrm{AST\ node}}\right)+0.3D_{\mathrm{AST\ depth}}$. We assign a larger weight to node count as it summarizes structural information over the complete syntax tree, whereas maximum depth depends only on the deepest root-to-leaf path and may be disproportionately affected by a single nested expression \citep{zuse2019software}. We parse Python programs using the built-in \texttt{ast} module. Matlab, R, and LaTeX programs are parsed using the corresponding language-specific parser.
    \item {\bf Plotting-Function Calls}: We also count function or method calls that directly affect graphical output. These include the creation of graphical primitives, axes, annotations, legends, color mappings, and figure layouts, but exclude utility functions unrelated to rendering. The plotting-operation feature is computed as $F_{\mathrm{API}}=\log\left(1+N_{\mathrm{API\ calls}}\right)$.
\end{itemize}
Finally, we perform equal-weighted averaging to compute the code complexity score, and the final difficulty score is $D_{\mathrm{overall}}=0.5D_{\mathrm{image}}+0.5D_{\mathrm{code}}$.
Finally, instances below the 33rd percentile of $D_{\mathrm{overall}}$ are assigned to the {\it Easy} tier, those between the 33rd and 67th percentiles to the {\it Medium} tier, and the remaining instances to the {\it Hard} tier. 
A valid ranking scheme should exhibit monotonically decreasing execution rates and perceptual similarity from the {\it Easy} to {\it Hard} tiers (Sec. \ref{exp::difficulty}). 
Fig. \ref{scatter_difficulty} shows the overall distributions and provides examples of image-code pairs at different difficulty tiers.

\subsection{Evaluation Protocol}
\label{subsec::taxonomy}
FigCodeBench incorporates multi-level evaluators that yield deterministic and reproducible assessment signals from both image-oriented and code-oriented perspectives by grounding evaluations in executable checks, quantitative, and qualitative measurements.

{\bf Image-Oriented Metrics.} 
To evaluate the visual equivalence between the rendered output of a generated script $\hat{I}$ and its ground-truth visualization $I$, we construct a multi-scale visual assessment suite that spans low-level pixel fidelity, structural layout integrity, and high-level human perception.
At the pixel and structural level, we employ Peak Signal-to-Noise Ratio (PSNR) to measure absolute reconstruction fidelity, and the Structural Similarity Index Measure (SSIM) \citep{wang2004image} to capture spatial alignment and structural continuity. To align with human perceptual judgment, we integrate Learned Perceptual Image Patch Similarity (LPIPS) \citep{zhang2018unreasonable} alongside the CIEDE2000 color difference formula ($\Delta E$) \citep{luo2001development}, which strictly quantifies global chromatic shifts in a perceptually uniform color space. Crucially, to prevent models from inflating visual metrics through a small sample of fortuitously compiled scripts, the final aggregate score is scaled as an execution-rate-weighted average, explicitly penalizing models with low compilation success rates.

{\bf Code-Oriented Metrics.}
In addition to the execution rate, we evaluate the generated code from lexical, structural, and plotting-semantic perspectives. Since visually equivalent figures can be produced by substantially different programs, no single reference-based code metric provides a complete characterization of code quality. 
We therefore adopt CodeBERTScore \citep{zhou2023codebertscore}, CrystalBLEU \citep{eghbali2022crystalbleu}, normalized AST similarity, and plotting-API F1 as complementary diagnostic metrics.
CodeBERTScore is used to measure the contextual semantic correspondence in Python only, due to the fact that it does not provide validated language-specific models for Matlab, R, or LaTeX.
Thus, we further compute CrystalBLEU for all four languages to quantify lexical correspondence, which excludes shared n-grams that occur frequently across unrelated programs, such as common import statements, figure-initialization routines, and standard rendering commands, preventing ubiquitous plotting templates from inflating similarity scores.
To measure structural correspondence, we compute a normalized AST similarity between the generated and ground-truth code. Specifically, we strip comments and formatting-related artifacts while preserving plotting-function names and arguments critical to visual rendering.
Given the normalized syntax trees \(T_g\) and \(T_r\) of the generated and reference code, respectively, we define
\begin{equation}
S_{\mathrm{AST}}=1-\frac{\operatorname{TED}(T_g,T_r)}{|T_g|+|T_r|},
\end{equation}
where \(\operatorname{TED}(\cdot,\cdot)\) denotes the unit-cost tree-edit distance and \(|T|\) is the number of nodes in tree \(T\). This normalization bounds the score between zero and one, with higher values indicating stronger structural similarity.
Unlike token-based metrics, normalized AST similarity captures correspondence in structures, nesting patterns, expressions, and function-call organization while remaining relatively insensitive to variable naming and formatting choices.
Finally, we introduce plotting-API F1 to measure whether the generated code invokes the plotting operations required to reconstruct the reference figure. 
Function calls are first mapped to a language-independent plotting ontology, such as LINE, SCATTER, BAR, AXIS\_CONFIG, LEGEND, ANNOTATION, SUBPLOT, and STYLE. 
For example, \texttt{matplotlib.pyplot.scatter}, \texttt{Matlab scatter}, \texttt{R geom\_point}, and the \texttt{TikZ/PGFPlots} option only marks are mapped to the canonical SCATTER operation. 
Let \(c_g(a)\) and \(c_r(a)\) denote the number of occurrences of canonical plotting operation \(a\) in the generated and reference code. Their multiset intersection is
\begin{equation}
    I_{\mathrm{API}}=\sum_{a}\min\!\left(c_g(a),c_r(a)\right).
\end{equation}
Plotting-API precision, recall, and F-score are then defined as
\begin{equation}
P_{\mathrm{API}}=\frac{I_{\mathrm{API}}}
{\sum_a c_g(a)},
\qquad
R_{\mathrm{API}}=
\frac{I_{\mathrm{API}}}
{\sum_a c_r(a)},
\qquad
F_{\mathrm{API}}
=
\frac{
2P_{\mathrm{API}}R_{\mathrm{API}}
}{
P_{\mathrm{API}}+R_{\mathrm{API}}
},
\end{equation}
which provides a task-specific and interpretable measure of whether the model recovered the essential graphical construction operations.

{\bf Mean Machine Opinion Score.}
Following the successful use of MLLMs for evaluation in text analysis and visual understanding tasks \citep{pmlr-v235-chen24h,pu2025judge,chen2025maceval,chen2025evaluating}, we employ the MLLM-as-a-Judge mechanism to evaluate the high-level logic. To avoid the bias caused by a single model's judgment, we refer to the subjective mean opinion score (MOS) in the field of image quality assessment \citep{chen2025study} and adopt the model jury form for rating.
Specifically, we input both the ground-truth figure and the generated figure into three MLLMs, and instruct them to output a high-level similarity score ranging from 0 to 100 based on the chart types, layout, text content, data, and style \citep{yangchartmimic}. After that, we average these scores across MLLM judges to get the overall mean machine opinion score (MMOS). 
The robustness demonstration of this rating strategy is provided in Sec. \ref{humaneval}.

{\bf The Figure-Code Fidelity Metric.}
The above unimodal metrics characterize the visual similarity of the rendered figures and the similarity of the generated code, respectively. Here, we further introduce Figure–Code Fidelity (FCF), a joint fidelity–efficiency metric that evaluates the visual quality achieved relative to the amount of code required. 
Let $Q_i\in[0,1]$ denote the composite visual-similarity score of the \(i\)-th generated figure, and let \(L_i^{\mathrm{gen}}\) and \(L_i^{\mathrm{gt}}\) denote the effective token counts of the generated and ground-truth code, respectively. We define the relative code-length ratio as $R_i^{\mathrm{len}}=L_i^{\mathrm{gen}}/L_i^{\mathrm{gt}}$ and the corresponding conciseness factor as 
\begin{equation}
    C_i^{\mathrm{len}}=
\frac{1}{\max(1,R_i^{\mathrm{len}})}.
\end{equation}
The instance-level FCF score is then defined as 
\begin{equation}
    \mathrm{FCF}_i=E_i Q_i C_i^{\mathrm{len}}=E_i\frac{Q_i}
{\max\left(1,L_i^{\mathrm{gen}}/L_i^{\mathrm{gt}}\right)},
\end{equation}
where $E_i$ is an execution indicator that equals one when the generated code is successfully executed and produces a valid figure, and zero otherwise. The reference-relative normalization controls for differences in intrinsic figure complexity and language-specific verbosity. Consequently, the FCF score remains bounded by the underlying visual-similarity score and reaches a high value only when the generated program is both visually faithful and concise.

Since the four image-oriented metrics operate on heterogeneous numerical scales, we transform each metric into a dimensionless similarity score for which a larger value consistently indicates greater visual fidelity. Specifically, we define the normalized PSNR similarity as 
\begin{equation}
    q_i^{\mathrm{PSNR}}=1-10^{-\mathrm{PSNR}_i/20}.
\end{equation}
For SSIM, we retain the original score without further transformation, while converting LPIPS into a bounded similarity score using an inverse transformation:
\begin{equation}
    q_i^{\mathrm{LPIPS}}=
\frac{1}{1+\mathrm{LPIPS}_i}.
\end{equation}
For the CIEDE2000 difference, we adopt the Just Noticeable Difference (JND)-based nonlinear mapping proposed by \citep{yang2012color}, where $\Delta E$ is converted into an objective score conforming to subjective color perception:
\begin{equation}
q_i^{\Delta E}=\frac{\mathcal{J}\left(\overline{\Delta E}_{i}\right)}{5}.
\end{equation}
At last, we aggregate them using an equally weighted geometric mean, preventing an exceptionally high score on one dimension from completely offsetting a substantial discrepancy on another:
\begin{equation}
\mathrm{Q}_i=\left(
q_i^{\mathrm{PSNR}}
q_i^{\mathrm{SSIM}}
q_i^{\mathrm{LPIPS}}
q_i^{\Delta E}
\right)^{1/4}.    
\end{equation}

\begin{table*}[t]
  \caption{Comparison of our FigCodeBench with other related datasets and benchmarks. ``I'', ``T'', and ``C'' denote the image, text, and code modalities, respectively. ``EM'' denotes the exact match. }
  \label{comparison}
  \centering
  \resizebox{1\linewidth}{!}{\begin{tabular}{lccccccc}
    \toprule
    {\bf Dataset/Benchmark} &{\bf Source}    & {\bf \# Lang.}  & {\bf \# Fig. Type}  & {\bf \# Test Inst.}&{\bf Reso.}&{\bf Eval. Format}&{\bf Metric} \\
    \midrule
    ChartLlama \citep{han2023chartllama}&Synthesized& 1 &10&458&704$\times$516&I+T$\rightarrow$T &GPT Score, EM  \\
     MMCode \citep{li2024mmcode}&Crawl&1&12& 263&534$\times$302&I+T$\rightarrow$C& Pass rate  \\
      MatPlotBench \citep{yang2024matplotagent} &Crawl&1&13&100&849$\times$639
&T$\rightarrow$C$\rightarrow$I& GPT Score  \\
      ChartX \citep{xia2025chartx}&Synthesized& 1&18&6,000& 1176$\times$819&I+T$\rightarrow$T\&C$\rightarrow$I&Multiple   \\
       Plot2Code \citep{wu2025plot2code}&Crawl&2&6&368&903$\times$664&I+T$\rightarrow$C$\rightarrow$I  &Multiple \\
        Design2Code \citep{si2025design2code}&Crawl&1&HTML&484& 1280$\times$1482&I+T$\rightarrow$C$\rightarrow$I& Multiple  \\
    ChartMimic \citep{yangchartmimic}&Crawl+Synthesized& 1&22 &2,400&1124$\times$782 & I+T$\rightarrow$C$\rightarrow$I &Multiple \\
    \midrule
   \rowcolor{gray!10} FigCodeBench (Ours)&Crawl+Synthesized&4&30&6,194&2022$\times$1552&I+T$\rightarrow$C$\rightarrow$I&Multiple\\
    \bottomrule
  \end{tabular}}
\end{table*}

\begin{figure*}[t]
\centerline{\includegraphics[width=1\textwidth]{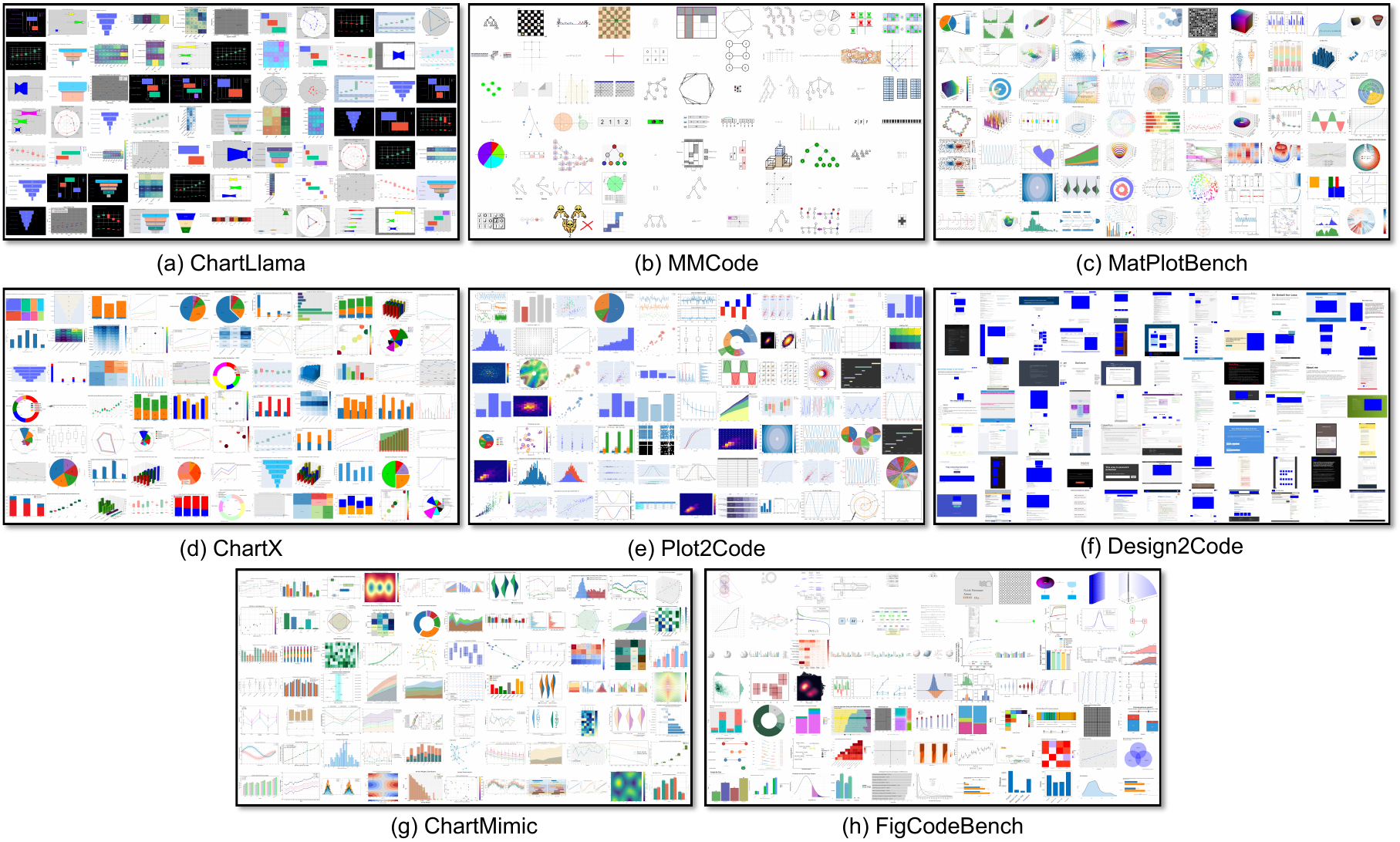}}
\caption{A gallery of random samples from seven compared benchmarks (i.e., ChartLlama, MMCode, MatPlotBench, ChartX, Plot2Code, Design2Code, and ChartMimic) and our proposed FigCodeBench. FigCodeBench covers 4 different languages, 7 categories, and 2,548 prototypes with 6,194 query images for standardized evaluation of MLLMs, significantly surpassing the existing benchmarks in terms of scale and coverage. Zoom-in for better visualization.}
\label{dataset_visualization}
\end{figure*}

\begin{figure*}[t]
\centerline{\includegraphics[width=1\textwidth]{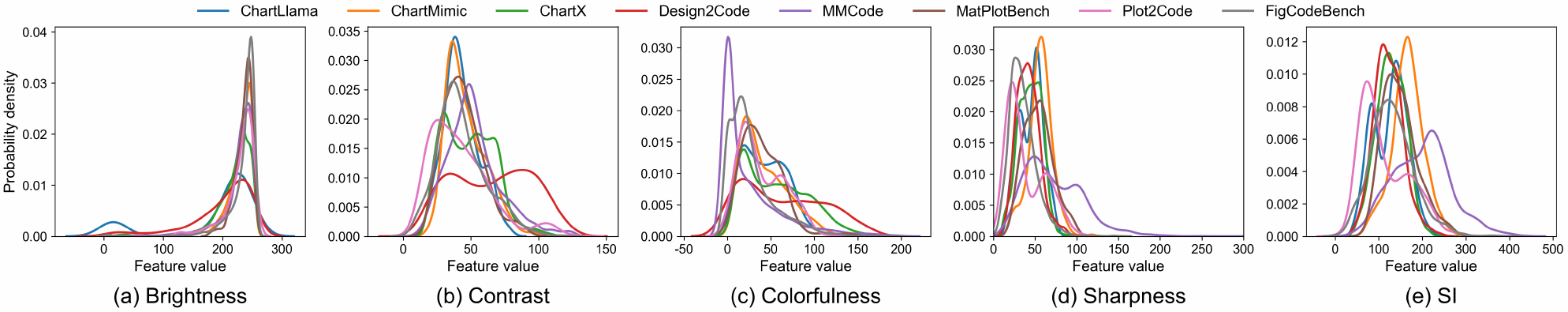}}
\caption{Feature distribution comparisons among eight benchmarks: ChartLlama, MMCode, MatPlotBench, ChartX, Plot2Code, Design2Code, ChartMimic, and our FigCodeBench.}
\label{feature_dist}
\end{figure*}

\begin{figure*}[t]
\centerline{\includegraphics[width=1\textwidth]{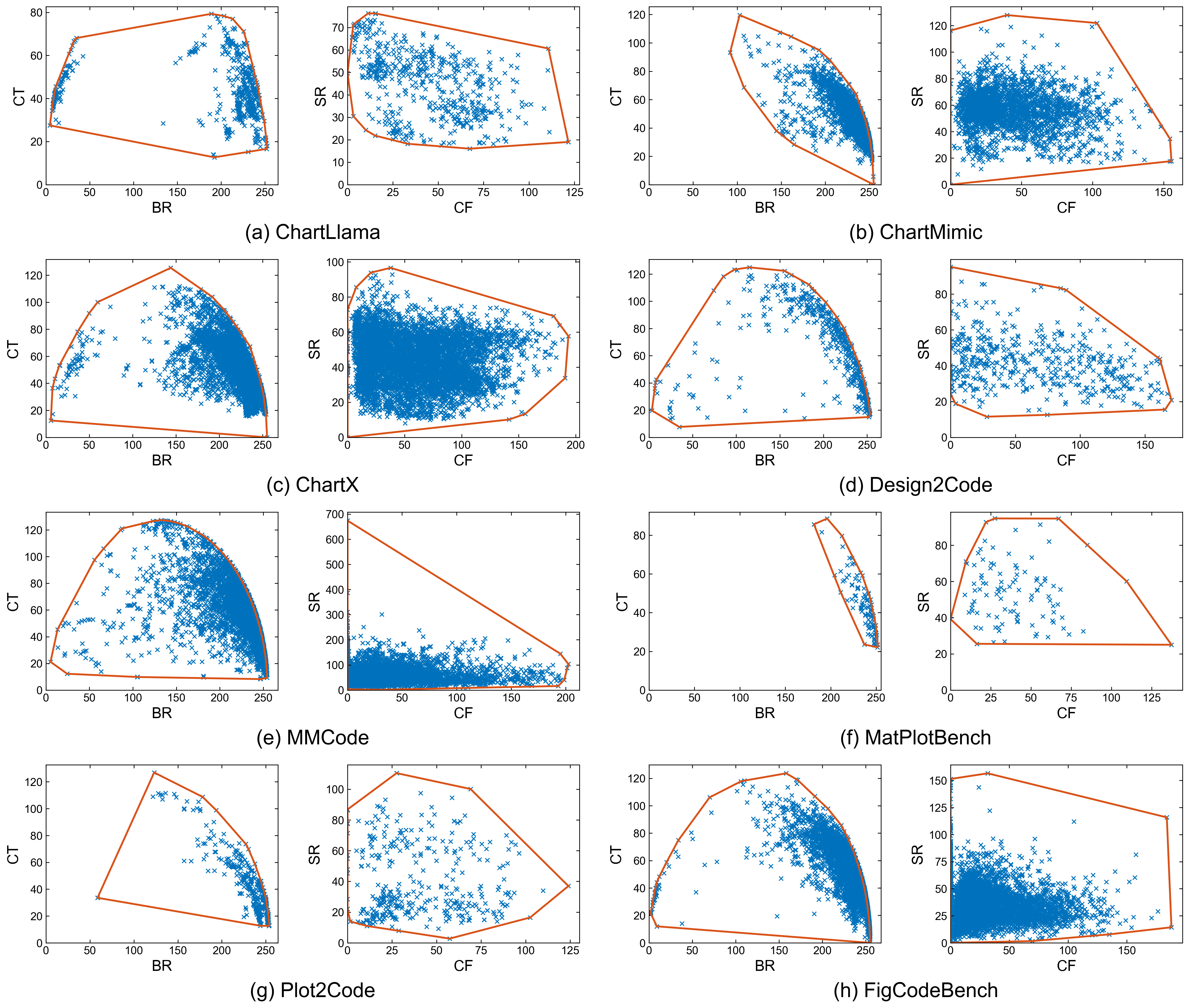}}
\caption{Query image (blue ‘x’) distribution in paired feature space with corresponding convex hulls (red boundaries). {\it Left}: Brightness (BR)$\times$Contrast (CT), {\it right}: Colorfulness (CF)$\times$Sharpness (SR).}
\label{convex_hull_dataset}
\end{figure*}

\subsection{Comparisons with Existing Benchmarks}
\label{subsec::comparisons}
To further distinguish the differences between FigCodeBench and the most related works \citep{han2023chartllama,xia2025chartx,li2024mmcode,yang2024matplotagent,wu2025plot2code,si2025design2code,yangchartmimic}, we summarize their main characteristics in Tab. \ref{comparison}. FigCodeBench distinguishes itself from the following perspectives: {\it (1) Real-world fidelity and less data contamination.} Unlike the synthesized data in ChartLlama \citep{han2023chartllama} and ChartX \citep{xia2025chartx} that may deviate from the distribution of real-world data, or raw crawled data in MMCode \citep{li2024mmcode} and Design2Code \citep{si2025design2code} that may suffer from data contamination, we not only enrich the data sources but also apply both LLM- and human-involved data augmentations to avoid these problems.
{\it (2) Covering more programming languages.} The existing chart-related code generation benchmarks mainly focus on Python, while we additionally include Matlab, R, and LaTeX languages. {\it (3) More diverse figure types.} FigCodeBench significantly surpasses existing benchmarks in both the diversity of figure types and the total data volume. {\it (4) Hybrid evaluators.} Recent benchmarks like ChartLlama \citep{han2023chartllama}, MatPlotBench \citep{yang2024matplotagent}, and ChartMimic \citep{yangchartmimic} rely heavily on GPTScore-based metrics \citep{fu2024gptscore}, which may raise concerns about the ranking robustness and model bias. 
In contrast, we aim to evaluate the figure reproduction capabilities of MLLMs by combining deterministic evaluators with an MLLM-as-Judge framework, including image-oriented and code-oriented metrics, to ensure comprehensiveness, accuracy, efficiency, and reproducibility.

\noindent
{\bf Content Diversity and Observations.}
To characterize the content diversity of the test images in each dataset, we include five low-level features \citep{tu2021ugc,chen2024gaia}, including brightness, contrast, colorfulness, sharpness, and spatial information (SI), thereby providing a larger visual space in which to plot and analyze content diversity of the eight benchmarks. Fig. \ref{dataset_visualization} provides a thumbnail of the compared benchmarks, and Fig. \ref{feature_dist} shows the fitted kernel distribution of each selected feature. We also plotted the convex hulls of paired features to show the feature coverage of each dataset in Fig. \ref{convex_hull_dataset}.
We make some observations from the above plots. As seen in Fig. \ref{dataset_visualization} and the corresponding convex hulls in Fig. \ref{convex_hull_dataset}, FigCodeBench and ChartX exhibit similar styles and coverage in brightness and contrast, while the styles of ChartLlama, MMCode, Design2Code, and MatPlotBench are relatively simple and their distribution is uneven.
On the sharpness and SI histograms, our FigCodeBench is spread most widely, while Plot2Code is concentrated on lower values. The overall distribution and coverage comparisons demonstrate that FigCodeBench possesses better diversity.

\begin{table*}[t]
    \centering
    \renewcommand\arraystretch{1}
    \caption{List of models evaluated and their respective organizations, cut-off date, and citations.}
    \resizebox{\linewidth}{!}{\begin{tabular}{llcc}
    \toprule                 
        {\bf Model Name}&{\bf Organizations}&{\bf Cut-off Date}&{\bf Citation}\\
        \midrule
        \texttt{gpt-5.4}& OpenAI&August 2025& \citep{gpt54}  \\
    \texttt{gpt-5.4-mini}& OpenAI&August 2025& \citep{gpt54}  \\
    \texttt{gpt-5.2}&OpenAI &August 2025& \citep{gpt52}  \\
    \texttt{claude-opus-4.7}&Anthropic &January 2026&  \citep{claude47opus} \\
    \texttt{gemini-3.1-pro-preview}&Google &January 2025& \citep{gemini31pro}  \\
    \texttt{gemini-3-flash-preview-nothinking}&Google &January 2025&\citep{gemini3flash}   \\
    \texttt{gemini-3-flash-preview-thinking}&Google &January 2025&\citep{gemini3flash}   \\
    \texttt{gemini-2.5-flash-nothinking}&Google &January 2025&  \citep{gemini25pro} \\
    \texttt{gemini-2.5-flash-thinking-16384}&Google &January 2025&  \citep{gemini25pro} \\
    \texttt{grok-4.3}&xAI &December 2025& \citep{grok43}  \\
    \texttt{qwen3.6-plus}&Alibaba &2026&  \citep{qwen36plus} \\
    \midrule
    \texttt{glm-5.1}&Zhipu &Not specified& \citep{zeng2026glm}  \\
    \texttt{glm-4.6v}&Zhipu &Not specified& \citep{hong2025glm}  \\
    \texttt{MiniMax-M3}&MiniMaxAI&January 2026&\citep{minimaxm3}\\
    \texttt{kimi-k2.5}&Moonshot AI &Not specified& \citep{team2026kimi}  \\
    \texttt{doubao-seed-2-0-pro-260215}&ByteDance&Not specified& \citep{seed20}  \\
    \texttt{qwen3.5-122b-a10b}&Alibaba &2026& \citep{qwen35}  \\
    \texttt{qwen3.5-35b-a3b}&Alibaba &2026& \citep{qwen35}  \\
    \texttt{qwen3-vl-235b-a22b-instruct}&Alibaba &March 2025& \citep{bai2025qwen3}  \\
    \texttt{qwen3-vl-30b-a3b-instruct}&Alibaba &March 2025& \citep{bai2025qwen3}  \\
    \texttt{qwen3-vl-8b-instruct}&Alibaba &March 2025&\citep{bai2025qwen3}   \\
    \texttt{qwen2.5-vl-72b-instruct}&Alibaba &June 2024&\citep{qwen25vl}   \\
    \texttt{internvl3.5-38b-instruct}&Shanghai AI Lab&Not specified&  \citep{wang2025internvl3_5}  \\
    \texttt{internvl3.5-8b-instruct}& Shanghai AI Lab&Not specified&\citep{wang2025internvl3_5}   \\
        \bottomrule
    \end{tabular}}
    \label{full_list}
\end{table*}

\begin{table*}[t]
  \caption{{\bf The evaluation resource consumption of FigCodeBench.} $AvgTok$ is the average number of tokens generated per query and $AvgCost$ is the approximate \$-cost per query.}
\renewcommand\arraystretch{1.05}
\belowrulesep=0pt\aboverulesep=0pt
  \label{resource_consumption}
  \centering
  \resizebox{1\linewidth}{!}{\begin{tabular}{l|cc|cc|cc|cc}
    \hline
    \multirow{2}{*}{\bf Model}     & \multicolumn{2}{c|}{\bf Python (n=987)}     & \multicolumn{2}{c|}{\bf Matlab (n=600)}&\multicolumn{2}{c|}{\bf R (n=1130)}&\multicolumn{2}{c}{\bf Latex (n=3477)} \\
    \cmidrule{2-3}\cmidrule{4-5}\cmidrule{6-7}\cmidrule{8-9}
&AvgTok&AvgCost&AvgTok&AvgCost&AvgTok&AvgCost&AvgTok&AvgCost\\
    \hline
    \multicolumn{9}{l}{\textit{Proprietary Models}} \\
    \hdashline
    GPT-5.4&814.98&\$0.0117&946.19&\$0.0122&898.51&\$0.0119&1176.65&\$0.0154    \\
    GPT-5.4-Mini&822.65&\$0.0034&702.30&\$0.0029&831.93&\$0.0040&1075.91&\$0.0043   \\
    GPT-5.2&792.12&\$0.0088&951.72&\$0.0110&830.21&\$0.0103&1026.63&\$0.0105   \\
    Claude Opus 4.7&597.10&\$0.0223&590.41&\$0.0204&341.74&\$0.0128&537.90&\$0.0192 \\
    Gemini 3.1 Pro&580.26&\$0.1222&619.17&\$0.0887&643.99&\$0.0899&861.01&\$0.1739 \\
    Gemini 3 Flash&558.61&\$0.0031&695.48&\$0.0030&629.90&\$0.0032&822.92&\$0.0035 \\
    Gemini 3 Flash-TK&640.40&\$0.0254&720.22&\$0.0226&655.11&\$0.0239&804.60&\$0.0323 \\
    Gemini 2.5 Flash&958.33&\$0.0019&1230.49&\$0.0019&765.14&\$0.0024&1236.69&\$0.0020 \\
    Gemini 2.5 Flash-TK&844.54&\$0.0200&1133.81&\$0.0191&912.07&\$0.0202&703.56&\$0.0209 \\
    Grok-4.3&644.26&\$0.0069&495.09&\$0.0059&572.09&\$0.0066&776.76&\$0.0052    \\
    Qwen3.6-Plus&686.35&\$0.0176&763.18&\$0.0217&660.09&\$0.0166&851.62&\$0.0108  \\
    \hline
    \multicolumn{9}{l}{\textit{Open-Source Models}} \\
    \hdashline
    GLM-5.1&1408.62&\$0.0244&1181.09&\$0.0168&774.71&\$0.0148&1080.14&\$0.0163 \\
    GLM-4.6V&497.02&\$0.0037&435.42&\$0.0028&386.01&\$0.0035&469.69&\$0.0037\\
    MiniMax-M3&860.15&\$0.0014&667.46&\$0.0031&637.89&\$0.0023&807.62&\$0.0049 \\
    Kimi-K2.5&698.12&\$0.0218&674.08&\$0.0220&644.64&\$0.0207&796.87&\$0.0172 \\
    DouBao-Seed-2-0-Pro&586.81&\$0.0069&459.35&\$0.0060&567.29&\$0.0084&716.40&\$0.0058 \\
    Qwen3.5-122B-A10B&529.24&\$0.0009&696.89&\$0.0010&577.12&\$0.0010&855.94&\$0.0012 \\
    Qwen3.5-35B-A3B&742.45&\textendash&875.25&\textendash&681.65&\textendash&841.01&\textendash \\
    Qwen3-VL-235B-A22B&475.23&\$0.0014&461.04&\$0.0012&477.29&\$0.0014&690.68&\$0.0013 \\
    Qwen3-VL-32B&968.89&\textendash&1670.08&\textendash&2043.48&\textendash&3440.89&\textendash  \\
    Qwen3-VL-8B&1309.17&\textendash&8301.24&\textendash&2820.2&\textendash&6753.37&\textendash \\
    Qwen2.5-VL-72B&366.18&\textendash&273.80&\textendash&379.81&\textendash&630.96&\textendash \\
    InternVL3.5-38B&535.13&\textendash&277.18&\textendash&428.75&\textendash&726.10&\textendash \\
    InternVL3.5-8B&1089.31&\textendash&2063.26&\textendash&2053.12&\textendash&2798.35&\textendash  \\
    \hline
  \end{tabular}}
\end{table*}

\section{Experiments}
\subsection{Evaluation Setup}

{\bf Evaluated Models.}
Our experiments include 24 MLLMs in total, with a mix of frontier proprietary models and open-source models. In particular, for proprietary models, we include models from OpenAI, Anthropic, Google, xAI, and Alibaba, e.g., \texttt{GPT-5.4} \citep{gpt54}, \texttt{GPT-5.2} \citep{gpt52}, \texttt{Claude Opus 4.7} \citep{claude47opus}, \texttt{Gemini 3.1 Pro} \citep{gemini31pro}, \texttt{Gemini 3 Flash} \citep{gemini3flash}, \texttt{Gemini 2.5 Pro} \citep{gemini25pro}, \texttt{Grok 4.3} \citep{grok43}, and \texttt{Qwen3.6-Plus} \citep{qwen36plus}. For open-source models, we include models from Zhipu, MiniMax, Moonshot AI, ByteDance, Alibaba, and Shanghai AI Lab, such as \texttt{GLM-5.1} \citep{zeng2026glm}, \texttt{MiniMax-M3} \citep{minimaxm3}, \texttt{Kimi-K2.5} \citep{team2026kimi}, \texttt{DouBao-Seed-2-0-Pro} \citep{seed20}, \texttt{Qwen3.5-122B-A10B} \citep{qwen35}, \texttt{Qwen3-VL-30B} \citep{bai2025qwen3}, and \texttt{InternVL3.5-38B} \citep{wang2025internvl3_5}. We provide detailed information with the corresponding versions, organizations, knowledge cut-off date, and citations for all participating models in Tab. \ref{full_list}.

{\bf Implementation Details.} 
Apart from the proprietary models and those with a parameter size greater than 100B that are deployed via API, all other models are run using up to 4 Nvidia H200 141GB GPUs. We use the default parameter settings (e.g., \texttt{temperature}, \texttt{top\_k}, and \texttt{top\_p}) of the respective models and do not apply an output token limit for better performance release. Regarding the generation errors caused by the absence of the package or library files, we have updated the local evaluation environment to prevent such occurrences. For GLM-5.1, although the official documentation does not explicitly state that it supports multimodal input, we use base64-encoded images as input.

{\bf Jury Models.} We employ three MLLMs to form the judging panel, i.e., gpt-5.1 \citep{gpt51}, gpt-5.6-luna  \citep{gpt56}, and gemini-3.5-flash \citep{gemini35}, which assess the extent to which the generated figure corresponds to the ground-truth figure while mitigating individual opinion bias. The specific prompt templates are presented in App. \ref{app::templates}.
Specifically, we input both the generated and the ground-truth figures into the judge models simultaneously. Then, each model is instructed to evaluate the similarity between the two figures, taking into account five dimensions, including text, layout, type, data, and style. Subsequently, each judge outputs an overall score ranging from 0 to 100 to represent the degree of similarity between the figures. Failed executions are directly assigned a score of 0.

\begin{table*}[!t]
  \caption{{\bf Image-oriented reproduction performance on FigCodeBench}. PSNR/SSIM ($\uparrow$) and LPIPS/$\Delta E$ ($\downarrow$) represent the pixel-level and perceptual metrics, respectively. ``TK'' denotes the {\it thinking} version. The best and second-best results are in {\bf bold} and \underline{underlined}.}
\renewcommand\arraystretch{1.15}
\belowrulesep=0pt\aboverulesep=0pt
  \label{metric::image}
  \centering
  \resizebox{1\linewidth}{!}{\begin{tabular}{l|crr|crr|crr|crr}
    \hline
    \multirow{2}{*}{\bf Model}     & \multicolumn{3}{c|}{\bf Python}     & \multicolumn{3}{c|}{\bf Matlab}&\multicolumn{3}{c|}{\bf R}&\multicolumn{3}{c}{\bf Latex} \\
    \cmidrule{2-4}\cmidrule{5-7}\cmidrule{8-10}\cmidrule{11-13}
&Exec.&PSNR/SSIM&LPIPS/$\Delta E$&Exec.&PSNR/SSIM&LPIPS/$\Delta \mathrm{E}$&Exec.&PSNR/SSIM&LPIPS/$\Delta E$&Exec.&PSNR/SSIM&LPIPS/$\Delta E$\\
    \hline
    \multicolumn{10}{l}{\textit{Proprietary Models}} \\
    \hdashline
    GPT-5.4&91.48 & \underline{12.807}/\underline{0.628} & \underline{0.411}/16.607&44.67&5.697/0.307&0.724/59.897&62.48&8.080/0.460&0.664/44.177&52.43&8.151/0.416 & 0.668/50.986    \\
    GPT-5.4-Mini&90.87&12.545/0.614&0.467/16.912&75.00&9.103/0.500&0.592/33.133&78.85&10.340/0.581& 0.571/28.265&48.12 & 7.764/0.388& 0.702/54.294   \\
    GPT-5.2&91.38 & 12.452/0.614& 0.487/\underline{16.459}&68.17& 8.239/0.446&0.649/39.135&41.33&5.248/0.297&0.786/62.414&39.69 & 6.218/0.315 & 0.760/62.473  \\
    Claude Opus 4.7&{\bf 99.29}&10.189/0.583 &  0.726/17.188&{\bf 98.17}&10.205/\underline{0.585}& 0.701/{\bf 15.603}& {\bf 97.35}&10.714/0.689&0.633/15.794&{\bf 88.47} & {\bf 12.822}/{\bf 0.683} & 0.594/{\bf 18.398}    \\
    Gemini 3.1 Pro&\underline{93.61}& {\bf 13.432}/{\bf 0.648} & {\bf 0.395}/{\bf 13.504}&82.50&10.414/0.562&\underline{0.495}/25.122&76.19&9.934/0.560& 0.587/30.874&59.76 & 9.772/0.486 & 0.610/43.148    \\
    Gemini 3 Flash&91.18& 12.702/0.619&0.445/16.472&81.00&9.939/0.544&0.536/26.808&90.44&\underline{12.040}/0.677&0.504/17.161&64.74 & 10.221/0.519 & \underline{0.593}/38.630  \\
    Gemini 3 Flash-TK&90.06& 12.495/0.608 & 0.459/17.254&\underline{88.33}&{\bf 11.115}/{\bf 0.595}&{\bf 0.479}/\underline{19.864}&\underline{93.81}&11.966/\underline{0.690}&0.515/{\bf 15.025}&60.22&9.511/0.484& 0.620/42.786   \\
    Gemini 2.5 Flash&87.93 & 9.128/0.545& 0.737/26.392&78.50&8.439/0.490&0.748/32.129&84.78&9.359/0.615&0.679/26.564&24.96 & 3.511/0.191& 0.884/76.987 \\
    Gemini 2.5 Flash-TK&81.03&8.789/0.502& 0.748/31.137&76.67&8.036/0.465&0.756/33.857&74.78&8.044/0.515&0.735/36.108&15.65 & 2.320/0.124& 0.921/85.388  \\
    Grok-4.3&90.97 & 11.528/0.583& 0.591/19.219&83.83&\underline{11.003}/0.381&0.605/25.803&85.49&11.569/0.646& 0.521/21.817&68.85 & 10.865/0.550 & 0.605/35.138 \\
    Qwen3.6-Plus&88.54 & 11.806/0.584 & 0.540/19.986&73.17&8.955/0.491&0.605/34.289&87.43&12.003/0.667&\underline{0.490}/19.679&60.05& 9.694/0.486& 0.630/43.067 \\
    \hline
    \multicolumn{10}{l}{\textit{Open-Source Models}} \\
    \hdashline
    GLM-5.1&83.67&9.037/0.497&0.754/29.444&64.33&6.550/0.364&0.813/45.116&64.16&6.873/0.431&0.770/45.551&40.41&5.839/0.311&  0.808/62.874 \\
    GLM-4.6V&86.11&10.580/0.568& 0.706/23.927&83.83& 8.956/0.470&0.753/27.531&86.55&10.844/0.615&0.668/22.373&23.12 & 3.332/0.179& 0.893/78.779  \\
    MiniMax-M3&88.34&9.125/0.514&0.748/27.301&77.50&7.283/0.451&0.773/36.782&76.73&8.778/0.525&0.721/33.622&63.19&8.556/0.479 & 0.722/42.355 \\
    Kimi-K2.5&92.29 & 12.297/0.614& 0.509/16.474&79.00&9.338/0.519&0.579/29.552&91.68&{\bf 12.610}/{\bf 0.699}&{\bf 0.459}/\underline{15.764}&48.75 & 7.846/0.393 &  0.707/53.895 \\
    DouBao-Seed-2-0-Pro&91.48 & 12.070/0.599 & 0.547/17.575&69.83&8.644/0.464&0.632/36.932&86.55&11.270/0.639& 0.527/21.257&49.01& 7.675/0.395&0.707/53.448 \\
    Qwen3.5-122B-A10B&82.56&10.635/0.563& 0.649/26.799&43.00&4.971/0.281&0.800/62.103&69.20&8.905/0.518&0.636/37.478&59.30&9.621/0.483 & 0.646/43.939 \\
    Qwen3.5-35B-A3B&70.18 & 9.236/0.460& 0.665/37.017&32.33&3.939/0.219&0.839/71.110&81.15&10.755/0.614&0.559/26.062&53.12& 8.699/0.436& 0.677/49.483  \\
    Qwen3-VL-235B-A22B&82.45&10.216/0.521 & 0.646/27.232&67.17&7.749/0.428&0.698/41.487&85.22&11.489/0.645&0.538/22.313&64.19 & 11.001/0.534 & 0.614/38.996 \\
    Qwen3-VL-32B&75.76&9.319/0.483& 0.677/33.363&40.33&4.429/0.245&0.824/65.247&69.20&9.066/0.519&0.636/37.212&53.98& 9.074/0.450&0.677/48.717 \\
    Qwen3-VL-8B&74.65 & 8.733/0.461 &  0.710/35.249&31.00&3.542/0.193&0.864/73.063&63.63&8.170/0.478&0.690/42.810&41.47& 6.732/0.341 & 0.752/60.976 \\
    Qwen2.5-VL-72B&79.51&9.152/0.506 & 0.670/31.460&59.17&6.654/0.379&0.721/48.506&81.15&10.912/0.619&0.574/26.688&\underline{75.87} & \underline{12.762}/\underline{0.633} & {\bf 0.556}/\underline{28.190} \\
    InternVL3.5-38B&77.69&9.242/0.492& 0.689/32.572&46.33&5.273/0.294&0.790/59.581&73.10&9.642/0.557&0.645/34.269&57.89&9.063/0.468&0.680/45.645 \\
    InternVL3.5-8B&65.42&7.483/0.409&0.747/43.868&37.83&4.246/0.240&0.831/67.195&56.90&7.402/0.435&0.730/49.011&43.20& 6.655/0.347 & 0.760/59.443  \\
    \hline
  \end{tabular}}
\end{table*}

\begin{table*}[!t]
  \caption{{\bf Code-level fidelity of figure reproduction performance on FigCodeBench.} We report CodeBERTScore (CBS; Python only), CrystalBLEU (CB), normalized AST similarity ($S_{\mathrm{AST}}$), and plotting-API F1 ($F_{\mathrm{API}}$) between the generated code and the corresponding ground-truth implementation. Higher values indicate greater similarity to the corresponding ground-truth code. Missing or unmatched code outputs are assigned a score of zero when computing the all-instance averages. The best and second-best results are in {\bf bold} and \underline{underlined}.}
\renewcommand\arraystretch{1.15}
\belowrulesep=0pt\aboverulesep=0pt
  \label{metric::code}
  \centering
  \resizebox{1\linewidth}{!}{\begin{tabular}{l|cccc|ccc|ccc|ccc}
    \hline
    \multirow{2}{*}{\bf Model}     & \multicolumn{4}{c|}{\bf Python}     & \multicolumn{3}{c|}{\bf Matlab}&\multicolumn{3}{c|}{\bf R}&\multicolumn{3}{c}{\bf Latex} \\
    \cmidrule{2-5}\cmidrule{6-8}\cmidrule{9-11}\cmidrule{12-14}
&CBS&CB&$S_{\mathrm{AST}}$&$F_{\mathrm{API}}$&CB&$S_{\mathrm{AST}}$&$F_{\mathrm{API}}$&CB&$S_{\mathrm{AST}}$&$F_{\mathrm{API}}$&CB&$S_{\mathrm{AST}}$&$F_{\mathrm{API}}$\\
    \hline
    \multicolumn{10}{l}{\textit{Proprietary Models}} \\
    \hdashline
    GPT-5.4&0.818&0.150&0.649&0.537&0.090&0.538&0.342&0.216&0.719&0.506&0.162&0.596&0.630\\
    GPT-5.4-Mini&0.812&0.138&0.630&0.518&0.111&0.590&0.415&0.215&0.725&0.494&0.152&0.599&0.617\\
    GPT-5.2&0.819&0.143&0.645&0.547&0.081&0.581&0.403&0.221&0.727&0.525&0.159&0.595&0.630\\
    Claude Opus 4.7&0.753&0.036&0.618&0.312&0.033&0.565&0.314&0.045&0.688&0.447&0.028&0.480&0.437\\
    Gemini 3.1 Pro&{\bf 0.845}&{\bf 0.274}&{\bf 0.741}&{\bf 0.646}&{\bf 0.136}&0.653&0.490&{\bf 0.306}&{\bf 0.797}&{\bf 0.591}&0.146&0.603&0.628\\
    Gemini 3 Flash&\underline{0.842}&\underline{0.265}&\underline{0.735}&\underline{0.605}&0.121&{\bf 0.669}&0.510&\underline{0.298}&\underline{0.787}&0.555&\underline{0.166}&{\bf 0.656}&{\bf 0.686}\\
    Gemini 3 Flash-TK&0.838&0.229&0.708&0.581&0.116&\underline{0.663}&0.507&0.288&0.773&0.542&0.134&0.611&0.648\\
    Gemini 2.5 Flash&0.752&0.027&0.559&0.254&0.020&0.550&0.297&0.063&0.676&0.377&0.050&0.456&0.354\\
    Gemini 2.5 Flash-TK&0.745&0.027&0.525&0.269&0.021&0.559&0.278&0.045&0.554&0.335&0.029&0.447&0.326\\
    Grok-4.3&0.819&0.180&0.680&0.533&0.111&0.595&0.501&0.263&0.786&\underline{0.570}&0.144&0.592&0.623\\
    Qwen3.6-Plus&0.824&0.190&0.669&0.536&0.105&0.638&0.476&0.273&0.772&0.534&{\bf 0.168}&\underline{0.645}&\underline{0.666}\\
    \hline
    \multicolumn{14}{l}{\textit{Open-Source Models}} \\
    \hdashline
    GLM-5.1&0.671&0.022&0.402&0.219&0.022&0.454&0.273&0.056&0.653&0.376&0.039&0.415&0.332\\
    GLM-4.6V&0.763&0.044&0.571&0.225&0.035&0.588&0.373&0.045&0.685&0.398&0.013&0.401&0.235\\
    MiniMax-M3&0.751&0.028&0.521&0.224&0.031&0.513&0.287&0.060&0.663&0.363&0.034&0.407&0.350\\
    Kimi-K2.5&0.816&0.192&0.658&0.538&\underline{0.130}&0.629&0.490&0.280&0.765&0.543&0.121&0.462&0.471\\
    DouBao-Seed-2-0-Pro&0.811&0.186&0.689&0.464&0.120&0.640&0.477&0.264&\underline{0.787}&0.544&0.124&0.539&0.572\\
    Qwen3.5-122B-A10B&0.815&0.200&0.698&0.557&0.104&0.640&0.481&0.252&0.773&0.542&0.155&0.643&0.649\\
    Qwen3.5-35B-A3B&0.797&0.151&0.608&0.482&0.084&0.630&0.453&0.245&0.758&0.520&0.158&0.613&0.648\\
    Qwen3-VL-235B-A22B&0.806&0.219&0.692&0.535&0.112&0.627&{\bf 0.525}&0.233&0.784&0.563&0.136&0.613&0.643\\
    Qwen3-VL-32B&0.817&0.136&0.663&0.484&0.042&0.590&0.504&0.086&0.721&0.459&0.069&0.582&0.604\\
    Qwen3-VL-8B&0.811&0.075&0.655&0.454&0.008&0.418&0.363&0.054&0.678&0.452&0.031&0.492&0.525\\
    Qwen2.5-VL-72B&0.812&0.177&0.666&0.502&0.063&0.536&\underline{0.516}&0.157&0.710&0.565&0.118&0.602&0.642\\
    InternVL3.5-38B&0.809&0.155&0.652&0.420&0.072&0.537&0.450&0.152&0.686&0.529&0.102&0.523&0.572\\
    InternVL3.5-8B&0.800&0.080&0.638&0.366&0.032&0.503&0.434&0.054&0.593&0.489&0.061&0.535&0.583\\
    \hline
  \end{tabular}}
\end{table*}

\begin{table*}[!t]
  \caption{{\bf Performance comparison on mean model opinion score (MMOS).} We report the scores rated by three MLLM judges. $\mathrm{J_1}$, $\mathrm{J_2}$, and $\mathrm{J_3}$ represent GPT-5.1, GPT-5.6-Luna, and Gemini-3.5-Flash, respectively.}
\renewcommand\arraystretch{1.15}
\belowrulesep=0pt\aboverulesep=0pt
  \label{metric::MMOS}
  \centering
  \resizebox{1\linewidth}{!}{\begin{tabular}{l|cccc|cccc|cccc|cccc}
    \hline
    \multirow{2}{*}{\bf Model}     & \multicolumn{4}{c|}{\bf Python}     & \multicolumn{4}{c|}{\bf Matlab}&\multicolumn{4}{c|}{\bf R}&\multicolumn{4}{c}{\bf Latex} \\
    \cmidrule{2-5}\cmidrule{6-8}\cmidrule{9-11}\cmidrule{12-17}
&$\mathrm{J_1}$&$\mathrm{J_2}$&$\mathrm{J_3}$&MMOS&$\mathrm{J_1}$&$\mathrm{J_2}$&$\mathrm{J_3}$&MMOS&$\mathrm{J_1}$&$\mathrm{J_2}$&$\mathrm{J_3}$&MMOS&$\mathrm{J_1}$&$\mathrm{J_2}$&$\mathrm{J_3}$&MMOS\\
    \hline
    \multicolumn{10}{l}{\textit{Proprietary Models}} \\
    \hdashline
    GPT-5.4 & 85.59 & 75.87 & \underline{77.74} & \underline{79.73} & 41.58 & 36.83 & 35.52 & 37.98 & 53.65 & 45.70 & 41.57 & 46.97 & 46.40 & 39.04 & 35.86 & 40.43 \\
    GPT-5.4-Mini & 84.31 & 74.60 & 66.76 & 75.22 & 65.70 & 56.62 & 56.64 & 59.65 & 67.46 & 57.88 & 51.12 & 58.82 & 41.27 & 33.86 & 30.57 & 35.23 \\
    GPT-5.2 & 85.43 & 75.52 & 68.04 & 76.33 & 61.77 & 53.87 & 52.06 & 55.90 & 34.97 & 29.59 & 26.43 & 30.33 & 35.86 & 29.77 & 26.97 & 30.87 \\
    Claude Opus 4.7&\underline{86.33}&\underline{77.59}&73.56&76.19&72.37&61.04&58.73&64.05&68.35&62.17&54.66&61.73&55.14&\underline{50.93}&{\bf 49.35}&\underline{51.81}\\
    Gemini 3.1 Pro & {\bf 88.20} & {\bf 80.66} & {\bf 82.24} & {\bf 83.70} & \underline{76.29} & \underline{67.98} & \underline{65.77} & \underline{70.01} & 65.48 & 56.51 & 51.74 & 57.91 & 56.24 & 50.44 & \underline{45.47} & 50.72 \\
    Gemini 3 Flash & 85.45 & 76.83 & 68.09 & 76.79 & 74.36 & 65.23 & 65.62 & 68.40 & 79.11 & \underline{69.17} & 61.00 & 69.76 & {\bf 59.96} & {\bf 52.28} & 45.04 & {\bf 52.43} \\
    Gemini 3 Flash-TK & 85.12 & 76.32 & 74.09 & 78.51 & {\bf 82.02} & {\bf 73.27} & {\bf 69.69} & {\bf 74.99} & {\bf 83.07} & 69.12 & {\bf 66.85} & {\bf 73.01} & \underline{56.62} & 49.93 & 42.78 & 49.78 \\
    Gemini 2.5 Flash&62.44&51.28&53.71&55.81&57.70&49.37&46.52&51.20&63.55&49.36&46.27&53.06&12.21&7.93&5.74&8.63\\
    Gemini 2.5 Flash-TK&63.79&54.06&58.68&58.84&53.66&47.68&45.80&49.05&58.21&51.60&50.32&53.38&8.94&6.58&3.87&6.46\\
    Grok-4.3 & 77.39 & 64.15 & 66.91 & 69.48 & 69.64 & 58.09 & 57.25 & 61.66 & 74.90 & 63.52 & 56.75 & 65.06 & 58.54 & 46.70 & 38.62 & 47.95 \\
    Qwen3.6-Plus & 80.89 & 70.33 & 70.58 & 73.93 & 65.35 & 57.36 & 55.71 & 59.47 & 74.08 & 65.14 & 63.36 & 67.53 & 54.65 & 46.67 & 43.45 & 48.26 \\
    \hline
    \multicolumn{14}{l}{\textit{Open-Source Models}} \\
    \hdashline
    GLM-5.1&47.86&41.83&42.92&44.20&27.84&19.23&21.28&22.78&52.77&38.33&32.64&41.25&27.65&18.35&16.77&20.92\\
    GLM-4.6V&53.44&47.92&49.31&50.22&34.11&29.73&29.70&31.18&74.13&58.83&48.22&60.39&12.47&7.31&6.85&8.88\\
    MiniMax-M3&58.60&51.64&54.74&54.99&29.05&23.18&25.63&25.95&57.28&43.37&37.29&45.98&45.76&37.57&32.60&38.64\\
    Kimi-K2.5 & 81.33 & 70.93 & 63.84 & 72.03 & 70.88 & 61.32 & 60.50 & 64.23 & \underline{82.60} & {\bf 72.31} & \underline{62.64} & \underline{72.52} & 43.87 & 36.60 & 32.12 & 37.53 \\
    DouBao-Seed-2-0-Pro & 82.27 & 70.20 & 71.00 & 74.49 & 62.02 & 53.23 & 50.90 & 55.38 & 76.33 & 64.48 & 58.18 & 66.33 & 44.22 & 37.31 & 35.29 & 38.94 \\
    Qwen3.5-122B-A10B & 42.58 & 35.48 & 39.27 & 39.11 & 32.02 & 26.40 & 27.52 & 28.65 & 51.60 & 43.12 & 40.98 & 45.23 & 49.61 & 40.90 & 37.36 & 42.62 \\
    Qwen3.5-35B-A3B & 61.10 & 52.52 & 53.81 & 55.81 & 26.73 & 22.34 & 21.79 & 23.62 & 63.64 & 54.59 & 51.90 & 56.71 & 46.64 & 38.69 & 34.49 & 39.94 \\
    Qwen3-VL-235B-A22B & 66.59 & 54.89 & 57.37 & 59.62 & 50.53 & 41.63 & 45.19 & 45.78 & 63.81 & 53.08 & 52.43 & 56.44 & 50.60 & 40.05 & 38.86 & 43.17 \\
    Qwen3-VL-32B & 59.19 & 48.11 & 49.77 & 52.36 & 29.08 & 23.77 & 25.12 & 25.99 & 50.01 & 41.33 & 40.35 & 43.90 & 43.21 & 34.38 & 32.53 & 36.71 \\
    Qwen3-VL-8B & 50.55 & 41.05 & 43.41 & 45.00 & 20.33 & 16.54 & 18.87 & 18.58 & 42.15 & 34.74 & 34.89 & 37.26 & 29.10 & 23.16 & 22.96 & 25.07 \\
    Qwen2.5-VL-72B & 46.31 & 38.62 & 43.61 & 42.85 & 41.62 & 33.65 & 35.98 & 37.08 & 55.17 & 46.04 & 47.00 & 49.40 & 53.12 & 41.28 & 41.05 & 45.15 \\
    InternVL3.5-38B & 50.27 & 41.30 & 43.80 & 45.12 & 30.51 & 25.54 & 27.92 & 27.99 & 38.87 & 32.33 & 36.51 & 35.90 & 36.74 & 28.67 & 30.15 & 31.85 \\
    InternVL3.5-8B & 38.66 & 31.67 & 34.79 & 35.04 & 23.41 & 19.13 & 21.09 & 21.21 & 27.95 & 23.70 & 27.33 & 26.33 & 24.22 & 19.56 & 21.15 & 21.64 \\
    \hline
  \end{tabular}}
\end{table*}

\subsection{Analysis of Evaluation Resource Consumption}
Tab. \ref{resource_consumption} reports the average resource consumption of 24 evaluated MLLMs in FigCodeBench. 
Based on this, we conduct a statistical analysis on the generation verbosity (AvgTok) and monetary overhead (AvgCost) to investigate whether the choice of programming language biases these overheads.

{\bf Setup.}
For the AvgTok analysis, all 24 evaluated MLLMs serve as independent subjects. For the AvgCost analysis, we evaluate 18 API-used models.
A preliminary normality check via the Shapiro-Wilk test indicates that both token and cost distributions are heavily right-skewed. This skewness is driven by two distinct factors: (1) severe length hallucinations in specific smaller-scale open-source models (e.g., Qwen3-VL-8B outputting over 8,300 tokens in MATLAB), and (2) orders-of-magnitude differences in vendor pricing paradigms (e.g., Gemini 3.1 Pro costing significantly more than standard models). To prevent these cross-model variances and outliers from violating parametric assumptions, we employ the non-parametric Friedman rank-sum test, which isolates the intra-model language differences by ranking the four languages within each model, mathematically neutralizing the confounding effect of varying baseline prices across vendors.

{\bf Results.}
As observed, Latex demands the highest token volume ($Median \approx 831$), sequentially followed by Matlab ($Median \approx 699$), Python ($Median \approx 692$), and R ($Median \approx 649$). The Friedman test confirms a statistically significant main effect of the programming language on the generated token length ($\chi^2(3, N=24) = 19.35, p < 0.001$). Subsequent pairwise post-hoc Nemenyi testing indicates that Latex exhibits a significantly higher AvgTok compared to both R ($p < 0.01$) and Python ($p < 0.05$) \citep{pereira2015overview}.
Meanwhile, the Friedman test on AvgCost reveals a significant main effect ($\chi^2(3, N=18) = 14.82, p < 0.01$). Ranking analysis confirms that LaTeX incurs the highest evaluation cost across the majority of models. For instance, the AvgCost for Gemini 3.1 Pro peaks at \$0.1739 per instance in Latex, representing an approximate 42\% cost inflation compared to its Python equivalent (\$0.1222). R emerges as the most economically efficient language to evaluate.
The significant divergence in evaluation resource consumption stems from the syntactic architecture of the respective languages. High-level data manipulation environments like Python and R leverage concise syntax, ensuring highly economical generation, while Latex (specifically TikZ/PGFPlots) functions as a declarative markup language, necessitating extensive boilerplate code and explicit coordinate-level geometric definitions.

\subsection{Benchmark Results}
\begin{figure*}[t!]
\centerline{\includegraphics[width=1\textwidth]{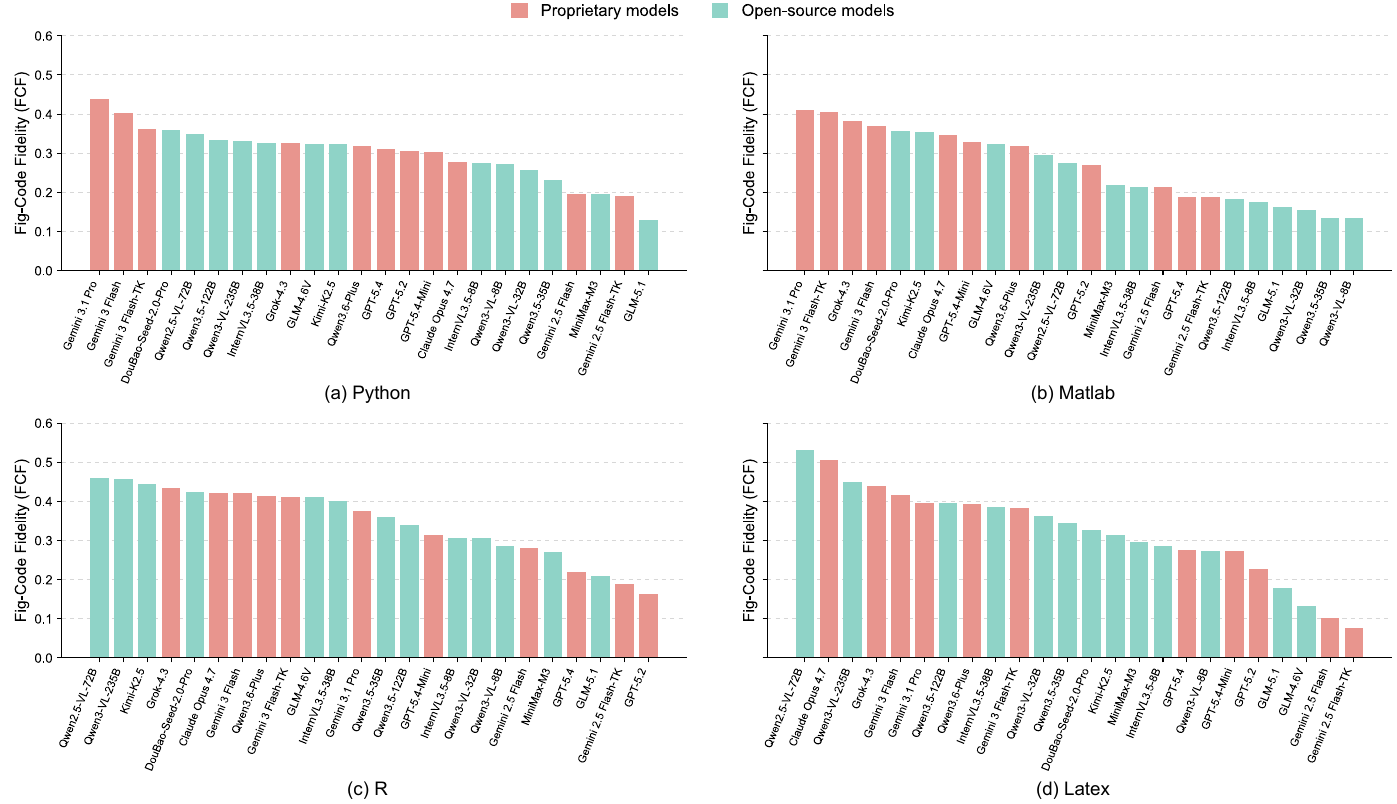}}
\caption{Comparison of Fig-Code Fidelity (FCF) results across four programming languages.}
\label{FCF}
\end{figure*}

Tab. \ref{metric::image}, Tab. \ref{metric::code}, Tab. \ref{metric::MMOS}, and Fig. \ref{FCF} show the main results in terms of image-oriented metrics, code-oriented metrics, mean model opinion score (MMOS), and fig-code fidelity (FCF) scores, respectively.

\begin{figure*}[t!]
\centerline{\includegraphics[width=1\textwidth]{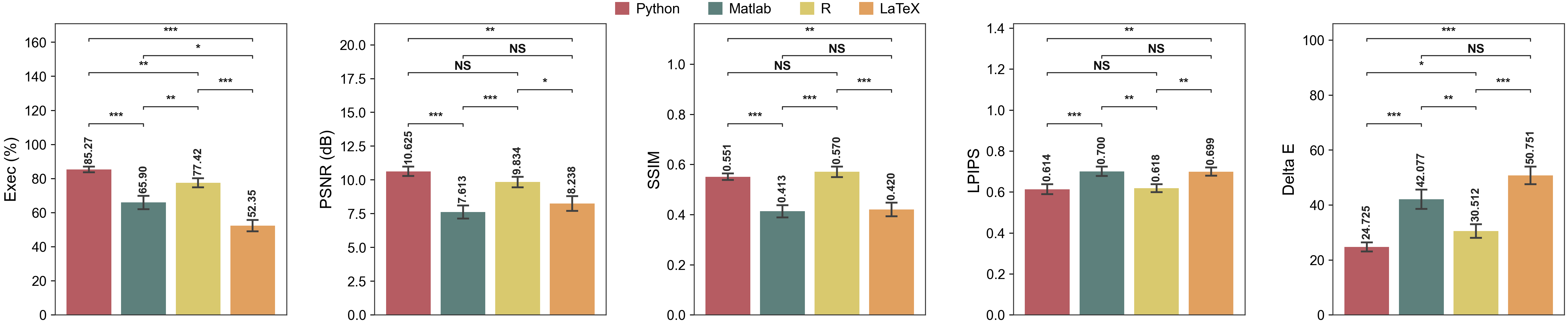}}
\caption{{\bf Statistical significance of cross-language differences in image-oriented performance.} Pairwise differences among Python, Matlab, R, and LaTeX are examined for execution success rate (Exec), PSNR, SSIM, LPIPS, and CIEDE2000 color difference (\(\Delta E\)). Bars and vertically displayed numbers show the mean performance across 24 evaluated models, and error bars denote the standard error of the mean. All $P$ values were calculated using two-sided Wilcoxon signed-rank tests on paired model-level measurements, followed by Holm–Bonferroni correction. NS, not significant; *\(\leq 0.05\), **\(\leq 0.01\), and ***\(\leq 0.001\).}
\label{statistical_test1}
\end{figure*}

{\bf Performance Comparison on Code Executability.} 
First, the results in Tab. \ref{metric::image} reveal a significant disparity in code executability. 
In Python evaluations, proprietary models show a highly concentrated, superior execution distribution ($Median = 90.97\%$), whereas open-source models exhibit lower and variable reliability ($Median = 82.45\%$), as supported by the Mann-Whitney U test ($U = 118.5, p < 0.01$). 
This gap further widens in less prevalent domains like Matlab, where proprietary models maintain a median execution rate of $78.50\%$, the open-source median collapses to $46.33\%$, yielding an even more pronounced statistical divergence ($U = 132.0, p < 0.001$). 
Fig. \ref{statistical_test1} further demonstrates that code executability is statistically biased by the target programming language, showing a performance hierarchy of $\text{Python} \succ \text{R} \succ \text{MATLAB} \succ \text{LaTeX}$. 
From the perspective of the model, Claude Opus 4.7 consistently achieves the highest execution rate across all 4 languages, followed by Gemini 3 Flash-TK and Grok-4.3. 
While flagship open-source models like Kimi-K2.5 and DouBao-Seed-2-0-Pro achieve highly competitive execution rates in Python ($>91\%$) and R ($>86\%$), their reliability also collapses when interacting with declarative markup (Latex) or domain-specific environments (Matlab).
Such execution degradation suggests that current pre-training paradigms may overweight Python data, leaving them poorly equipped to generate executable code for other language ecosystems.
Additionally, a longitudinal comparison between the Gemini 2.5 and 3 series highlights a significant generational leap in cross-domain compilation resilience, with execution gains observed across all four languages, including an improvement of nearly 40\% in execution rate for Latex.

{\bf Performance Comparison on Image-oriented Metrics.}
From the perspective of image-oriented reproduction metrics, first, high code execution reliability does not necessarily translate into superior image reproduction quality, indicating that successful code generation and visual rendering constitute related but distinct capabilities.
Specifically, while Claude Opus 4.7 achieves a state-of-the-art execution rate in Python tasks, it is substantially outperformed by Gemini 3.1 Pro and GPT-5.4 across all visual assessment criteria. In both Matlab and R languages, Claude Opus 4.7 maintains commanding execution rates ($>97\%$) but is consistently eclipsed in visual similarity by the Gemini 3 Flash series.
This phenomenon highlights a trade-off in code generation strategies, i.e., compilation conservatism versus visual risk-taking. Models like Claude Opus 4.7 appear to prioritize syntactic safety, thereby guaranteeing compilation success but resulting in generic charts that fail to capture the fine-grained aesthetic intricacies of the ground truth.
Second, in languages like Python and R, models generally achieve high SSIM ($>0.550$) and low LPIPS ($<0.600$), suggesting that the underlying coordinate mapping and structural layout algorithms are well-learned. However, transitioning to Matlab and Latex triggers a severe collapse in visual fidelity. 
As demonstrated in Fig. \ref{statistical_test1}, this degradation is most apparent in the $\Delta E$ metric. For instance, while GPT-5.4 maintains a $\Delta E$ of $16.607$ in Python, its perceptual color error increases to $59.897$ in MATLAB and $50.986$ in Latex, indicating that even when models successfully compile low-resource code, they often fall back on hallucinated color schemes, failing to achieve the strict pixel-level determinism required for scientific reproduction.
Furthermore, while the open-source MLLMs generally lag behind proprietary models, remarkable domain-specific breakthroughs by leading open-source architectures are also observed. 
Notably, Kimi-K2.5 sets a new benchmark record in the R domain. It not only achieves a formidable execution rate ($91.68\%$) but also entirely dominates the image-oriented metrics with the highest PSNR ($12.610$), SSIM ($0.699$), and lowest LPIPS ($0.459$) across all 24 evaluated models.

\begin{figure*}[t!]
\centerline{\includegraphics[width=1\textwidth]{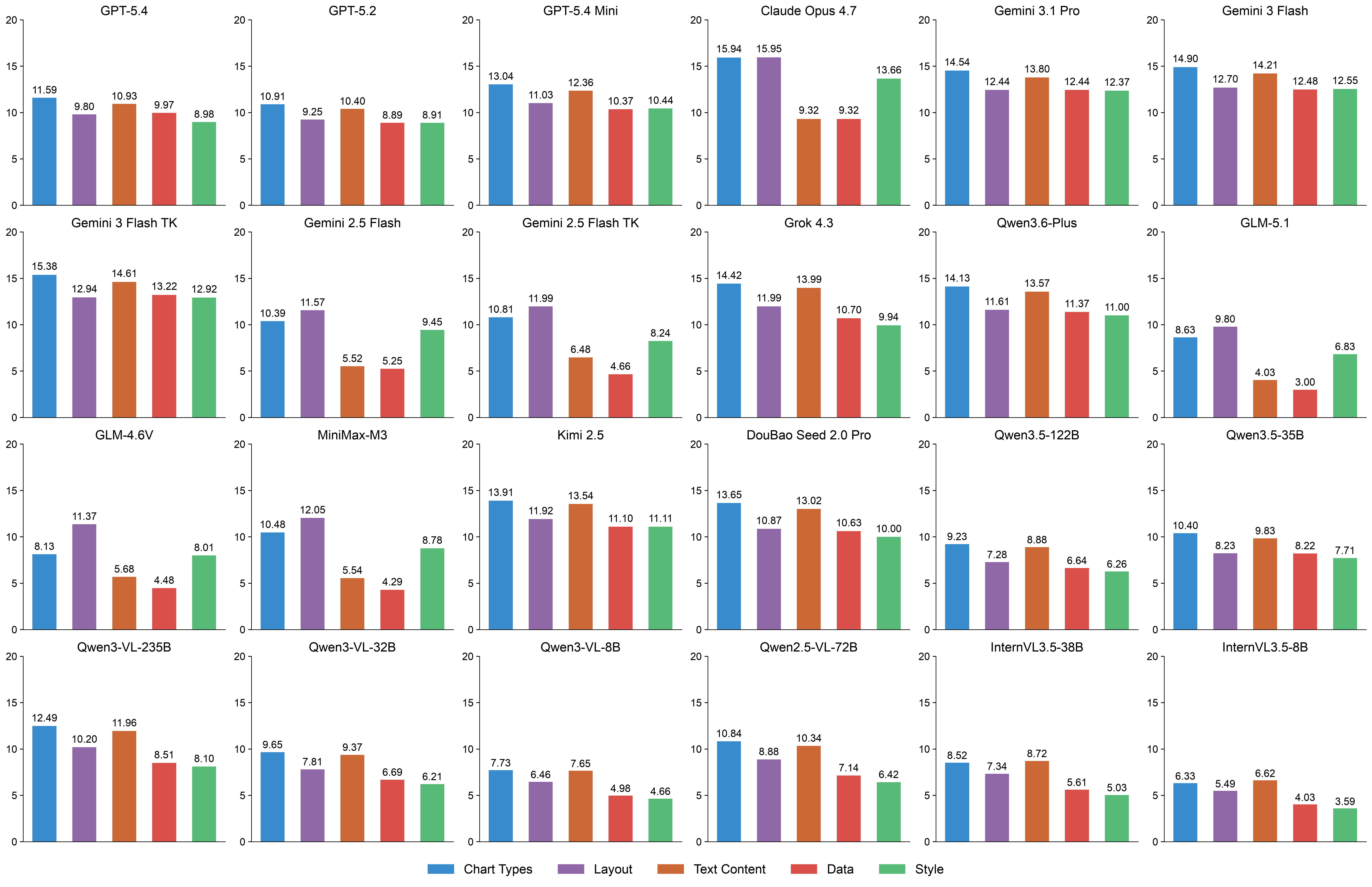}}
\caption{{\bf Fine-grained score distribution generated by MLLM judges.} We report the average scores of three MLLM judges for each evaluated model across four programming languages.}
\label{attribute-wise-mllm-judge}
\end{figure*}

\begin{figure}[t!]
\centerline{\includegraphics[width=.7\textwidth]{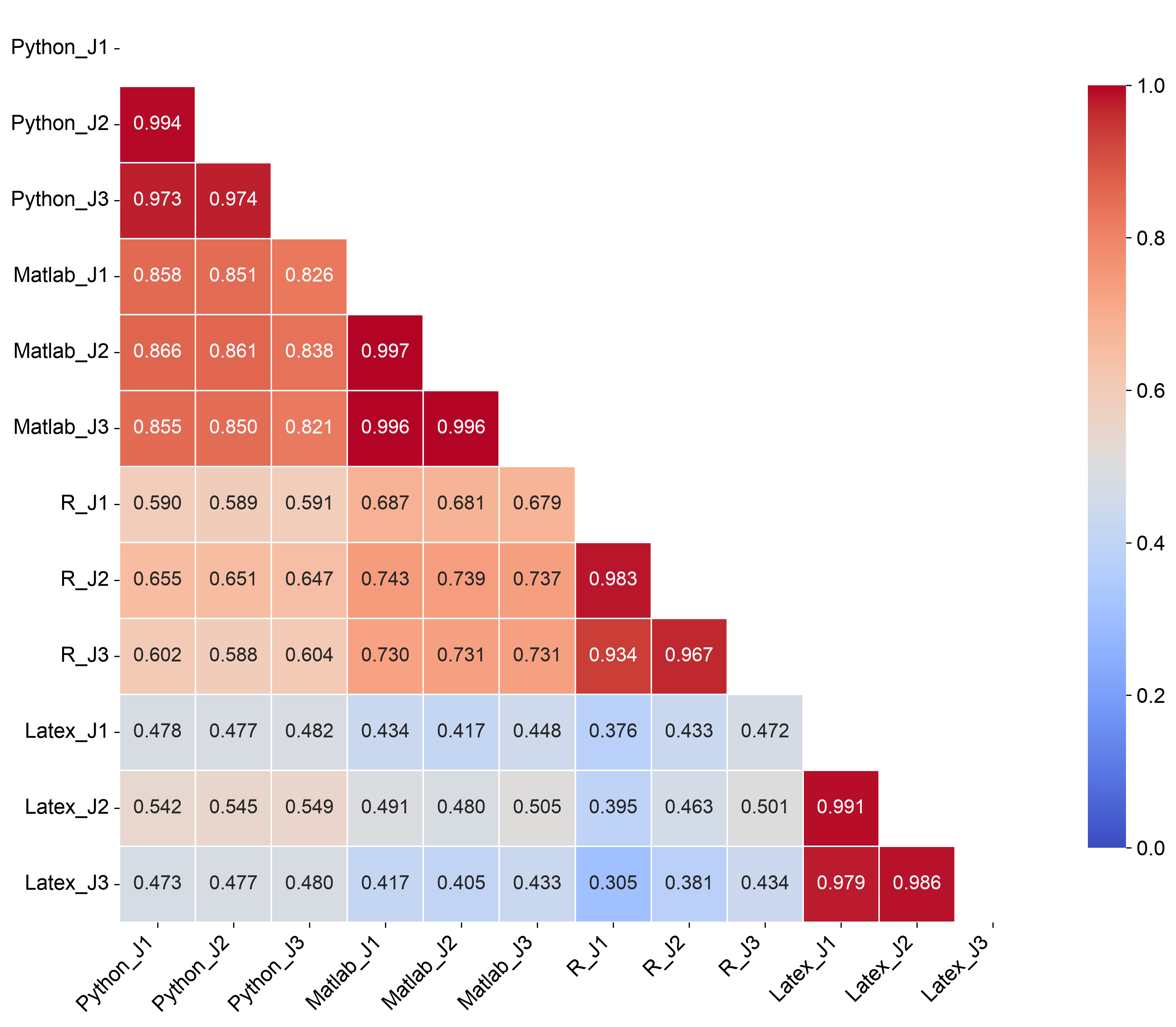}}
\caption{Pearson correlation between different MLLM judges across four programming languages.}
\label{corr::inMLLMjudge}
\end{figure}

{\bf Performance Comparison on Code-oriented Metrics.}
We quantify the syntactic and structural similarity of the generated code against the ground-truth scripts in Tab. \ref{metric::code}.
Surprisingly, Claude Opus 4.7, despite achieving the state-of-the-art in execution reliability ($>98\%$) and image-oriented fidelity across all domains, exhibits near-bottom code similarity scores.
In Python, its CrystalBLEU is only 0.036, and its API F1-score is 0.312, substantially lower than comparable models such as Gemini 3.1 Pro and even lower than much smaller models like Qwen3-VL-8B.
This indicates that Claude Opus 4.7 does not memorize or mimic the canonical, textbook-style snippets in the ground-truth dataset. Instead, it adopts alternative programmatic paradigms that are structurally divergent (low $S_{\mathrm{AST}}$) but visually isomorphic and computationally robust.
Meanwhile, models that perform well on image metrics unexpectedly also achieve high scores on these metrics. 
For example, Gemini 3.1 Pro achieves the highest $S_{\mathrm{AST}}$ ($0.741$) and $F_{\mathrm{API}}$ ($0.646$) in Python, followed by Gemini 3 Flash. A similar pattern holds in Matlab and R, exhibiting a strong tendency to reproduce canonical syntax from human-written code. 
Moreover, a cross-metric comparison reveals that exact lexical matching (CrystalBLEU) yields uniformly low scores across all models and domains (peaking at only $0.306$ for Gemini 3.1 Pro in R). This highlights the inferiority of token-level evaluation for code generation, as variable renaming or superficial formatting changes heavily penalize the score. In contrast, $S_{\mathrm{AST}}$ and $F_{\mathrm{API}}$ yield significantly higher and more discriminative distributions, confirming that while models rarely guess the exact tokens of the ground truth, they successfully capture the essential plotting functions required for visual synthesis.

\begin{figure*}[t!]
\centerline{\includegraphics[width=1\textwidth]{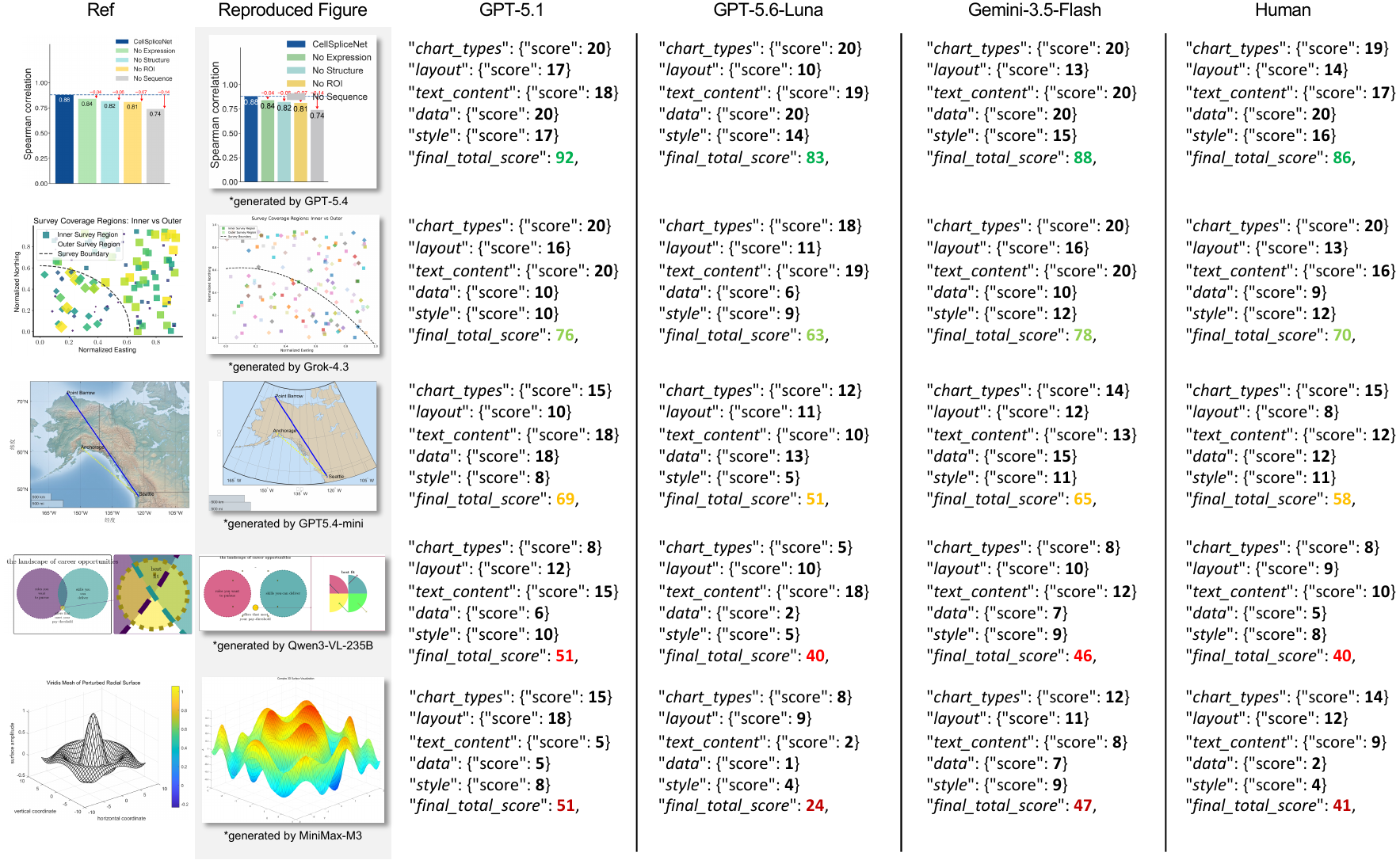}}
\caption{{\bf Qualitative visualization of MLLM judges.} All judge models produce human-aligned scores and mutually consistent rankings. From top to bottom, the reproduced figures show progressively decreasing similarity to the reference.}
\label{judge-example}
\end{figure*}

{\bf MLLM Judgment.}
As reported in Tab. \ref{metric::MMOS}, Gemini 3.1 Pro and Gemini 3 Flash series achieve the highest MMOS in Python and the other environments, respectively, which marks a substantial performance leap over their predecessor Gemini 2.5. 
Although open-source MLLMs generally lag behind proprietary ones, a few frontier models achieve remarkable domain-specific alignment.
Specifically, Kimi-K2.5 achieves an exceptional MMOS of 72.52 in R, significantly outperforming numerous peers and even surpassing top-tier proprietary models like Claude Opus 4.7 (61.73). DouBao-Seed-2-0-Pro similarly demonstrates highly competitive visual fidelity in Python (74.49). 
However, most open-source models (e.g., the GLM and smaller InternVL families) exhibit catastrophic performance degradation on Matlab and Latex tasks, with scores frequently falling into the 20–30 range. 
This extreme variance highlights a severe skew in open-source instruction-tuning distributions, indicating a critical scarcity of syntactic and visual priors in long-tail, low-resource plotting ecosystems. 
Consistent with prior multi-dimensional evaluations, rendering via Latex constitutes a formidable bottleneck for all evaluated models.
The highest MMOS values, achieved by Gemini 3 Flash (52.43) and Claude Opus 4.7 (51.81), barely cross the threshold of marginal acceptability, underscoring that current MLLMs, regardless of parameter scale or architectural openness, have yet to master the explicit mathematical coordinate mapping, precise node anchoring, and syntactic invariants demanded by low-level declarative frameworks like TikZ and PGFPlots.

Fig. \ref{attribute-wise-mllm-judge} further provides the rating results for five sub-dimensions, i.e., chart types, layout, text content, data, and style. We can observe that across nearly all evaluated models, the {\it chart types} (blue) and {\it layout} (purple) dimensions have the highest scores, while {\it data} (red) consistently performs as the weakest performance bottleneck. 
This indicates that MLLMs are generally proficient at identifying the correct base plotting API while suffering a profound degradation when tasked with microscopic precision, specifically the accurate mathematical projection of data distributions and geometric mappings into the coordinate space. 
Such discrepancy is magnified in mid-tier and earlier-generation models, such as Claude Opus 4.7, Gemini 2.5 Flash, and GLM-4.6V, which maintain competent layout scores but experience a significant performance decrease in both text content and data.
In contrast, models like Gemini 3 Flash-TK that achieve the highest overall MMOS also demonstrate a balanced integration across all five axes.
A similarly balanced capability profile is mirrored by top-tier open-source models like Kimi 2.5 and DouBao Seed 2.0 Pro, indicating that frontier multimodal alignment techniques are successfully bridging the gap between simply writing executable code and orchestrating a deterministic, pixel-perfect visual rendering engine.

Furthermore, we observe slight deviations in evaluation preference. GPT-5.1 ($J_1$) exhibits the most lenient scoring bias, awarding the highest scores across all model categories. In contrast, GPT-5.6-Luna ($J_2$) and Gemini-3.5-Flash ($J_3$) exhibit markedly more conservative and rigorous baselines. This intrinsic score variance strongly validates the necessity of the MMOS formulation, which calibrates individual judge leniency to yield a reliable and robust metric for visual generation fidelity.   
As shown in Fig. \ref{corr::inMLLMjudge}, all three MLLM judges correlate positively with each other within all languages, demonstrating the feasibility of our rating strategy. Meanwhile, the low correlations in cross-language scoring reflect domain brittleness and a severe divergence in the models' capabilities.
Fig. \ref{judge-example} further illustrates some qualitative results of MLLM judges, showing that their assessments are largely consistent across the five evaluation aspects.

\begin{table*}[!t]
  \caption{{\bf Performance comparison across different functional categories of figures}. MMOS is used as the indicator. }
\renewcommand\arraystretch{1.15}
\belowrulesep=0pt\aboverulesep=0pt
  \label{metric::divisions}
  \centering
  \resizebox{1\linewidth}{!}{\begin{tabular}{l|ccccccc}
    \hline
    {\bf Model}& {\bf Statistical} & {\bf Relational }&{\bf Temporal}&{\bf Compositional}&{\bf Geospatial}&{\bf Mathematical}&{\bf Conceptual} \\
    \hline
    \multicolumn{6}{l}{\textit{Proprietary Models}} \\
    \hdashline
    GPT-5.4&61.15&55.01&54.30&55.16&41.08&56.56&40.43\\
    GPT-5.4-Mini&69.28&61.66&65.81&59.63&36.47&66.02&35.23\\
    GPT-5.2&52.19&51.09&55.25&43.77&21.67&60.06&30.87\\
    Claude Opus 4.7&68.76&62.31&72.58&66.74&30.75&61.48&\underline{51.81}\\
    Gemini 3.1 Pro&70.84&66.92&72.42&64.00&{\bf 65.57}&\underline{72.50}&50.72\\
    Gemini 3 Flash&75.31&\underline{69.82}&74.60&\underline{74.55}&\underline{53.90}&67.56&{\bf 52.43}\\
    Gemini 3 Flash-TK&{\bf 78.33}&{\bf 72.44}&{\bf 79.60}&{\bf 74.59}&49.08&{\bf 74.76}&49.78\\
    Gemini 2.5 Flash&52.39&47.41&54.61&42.98&26.57&45.28&8.63\\
    Gemini 2.5 Flash-TK&54.36&47.30&56.27&42.56&24.36&48.55&6.46\\
    Grok-4.3&71.45&62.06&68.43&65.54&41.62&61.56&47.95\\
    Qwen3.6-Plus&74.49&62.20&72.28&64.38&35.56&65.20&48.26\\
    \hline
    \multicolumn{6}{l}{\textit{Open-Source Models}} \\
    \hdashline
    GLM-5.1&42.37&35.18&48.94&32.84&19.28&41.25&20.92\\
    GLM-4.6V&46.89&40.43&54.67&35.14&24.35&39.89&8.88\\
    MiniMax-M3&51.50&45.66&52.92&38.60&29.07&46.74&38.64\\
    Kimi-K2.5&\underline{76.05}&66.67&\underline{75.92}&71.74&38.86&63.77&37.53\\
    DouBao-Seed-2-0-Pro&72.01&63.28&72.44&71.33&37.99&58.68&38.94\\
    Qwen3.5-122B-A10B&44.38&33.99&40.23&42.52&17.82&36.76&42.62\\
    Qwen3.5-35B-A3B&55.95&46.80&53.59&39.00&21.98&43.95&39.94\\
    Qwen3-VL-235B-A22B&59.59&52.17&57.30&46.93&35.36&55.41&43.17\\
    Qwen3-VL-32B&49.53&37.07&43.75&43.14&16.93&40.61&36.71\\
    Qwen3-VL-8B&40.55&29.51&37.97&31.76&16.68&37.15&25.07\\
    Qwen2.5-VL-72B&49.89&38.45&43.71&46.71&25.48&42.61&45.15\\
    InternVL3.5-38B&40.62&32.37&35.80&33.64&21.83&42.40&31.85\\
    InternVL3.5-8B&33.29&20.14&27.49&28.03&4.40&32.01&21.64\\
    \hline
  \end{tabular}}
\end{table*}

{\bf Performance Comparison on Generation Efficiency.}
While image-oriented metrics evaluate the absolute visual quality and code-oriented metrics assess syntactic alignment, true practical utility in code generation hinges on token economy. To evaluate this, we analyze the Figure–Code Fidelity (FCF), a joint metric that quantifies the visual quality achieved relative to the volume of code generated. The FCF distributions across the four programming domains, visualized in Fig. \ref{FCF}, reveal a highly asymmetric landscape of generative efficiency.
In the Python and Matlab tasks, proprietary models establish a clear Pareto frontier for generation efficiency. Gemini 3.1 Pro consistently achieves the highest FCF scores in both domains, followed closely by the Gemini 3 Flash series. This indicates that these proprietary architectures have successfully internalized the canonical, high-density APIs of libraries like matplotlib and base Matlab, allowing them to achieve maximum visual fidelity without resorting to verbose workarounds.

Conversely, transitioning to R and Latex reveals a striking inversion of the performance hierarchy. 
In these domains, the open-source MLLMs outperform proprietary models in generative efficiency. Most notably, Qwen2.5-VL-72B obtains the highest FCF scores in both R and Latex, while GPT-5.2 and Gemini 2.5 Flash are in last place.
Given the high degree of similarity in their code structures (Tab. \ref{metric::code}) and the relatively smaller number of generated tokens (Tab. \ref{resource_consumption}), we speculate that these models execute complex geometric mapping with concise macro commands, whereas many proprietary models suffer from length penalties by over-generating explicit coordinate logic.

{\bf Performance on Different Image Types.}
In Tab. \ref{metric::divisions}, we investigate the figure reproduction performance of models across seven functional categories. 
First, most models show the highest proficiency in charting tasks that rely on standard data mapping, namely the statistical, temporal, and compositional categories. Notably, Gemini 3 Flash-TK, augmented with test-time reasoning, dominates, achieving the highest scores of 78.33, 79.60, and 74.59 across all three categories, respectively.
Second, the geospatial category constitutes the most significant obstacle in current visual code generation tasks. 
Except for Gemini 3.1 Pro, which achieves a relatively high score of 65.57, almost all other models suffer a cliff-like performance collapse. For instance, GPT-5.2 obtains a mere 21.67, while InternVL3.5-8B reaches only 4.40. This failure can be attributed to the underlying logic of map rendering, which requires models to process non-standard map projections and frequently relies on implicit, external geographic polygon data (such as Shapefiles or specific map packages). Without explicit definitions of these complex coordinate contours in their pre-training corpora, models can produce severe logical hallucinations when generating such code.  
Furthermore, the conceptual and relational categories, which emphasize abstract logic and relationships, also exhibit severe capability degradation.
This further exposes a severe architectural flaw, i.e., current MLLMs lack the 2D spatial reasoning capabilities required for code-based object depiction and the representation of abstract set relationships (e.g., Venn diagrams or hierarchical trees), once detached from a regular grid coordinate system.

\begin{figure*}[t!]
\centerline{\includegraphics[width=1\textwidth]{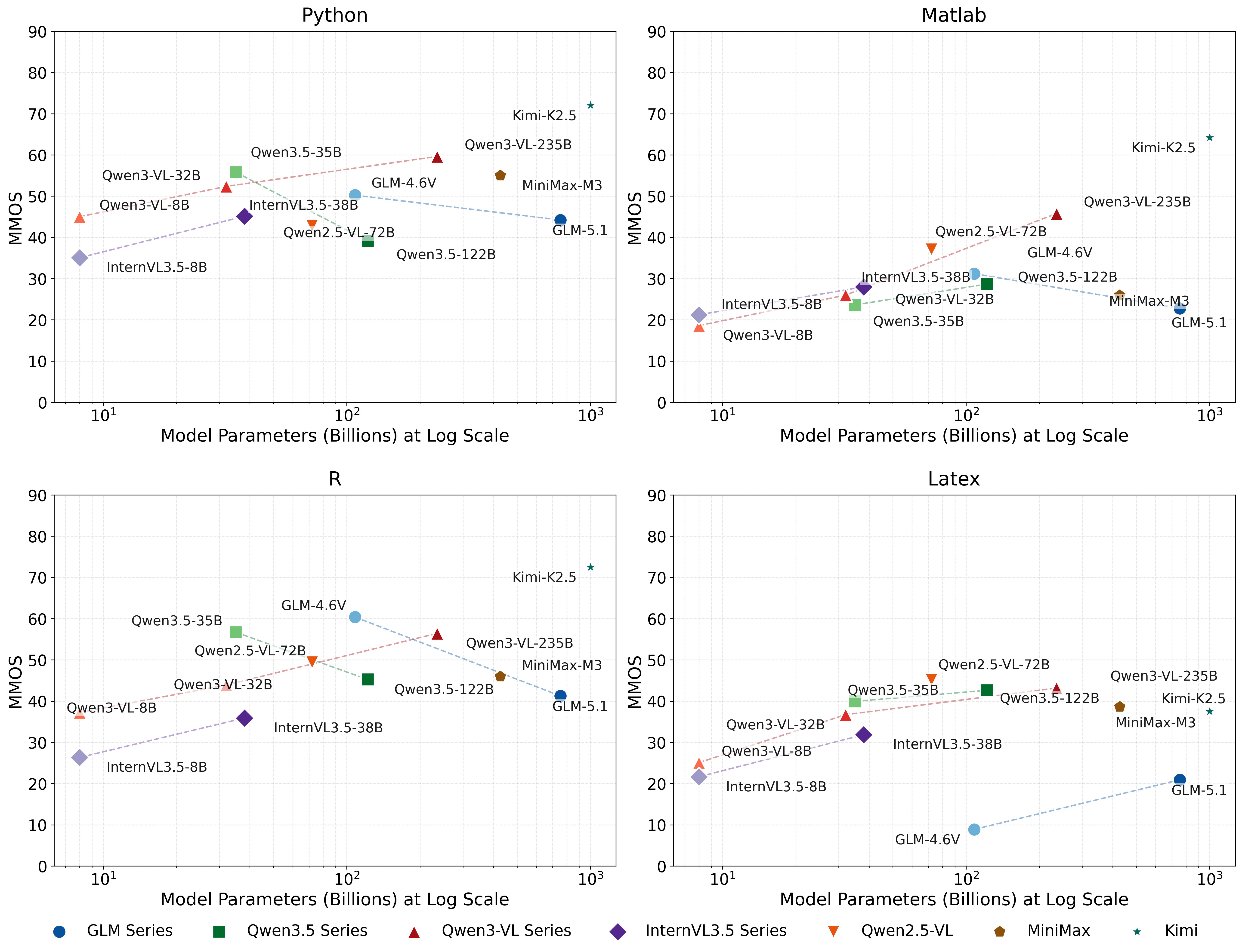}}
\caption{{\bf Model size versus MMOS}. Trend lines are added to show the performance trend of each model family across different parameter scales. Note that we only investigate open-source MLLMs with publicly available parameters.}
\label{modelsize}
\end{figure*}

{\bf Comparison Between Reasoning Models and Their Non-reasoning Counterparts.}
We also examine the impact of enabling reasoning in MLLMs on different languages. In particular, we choose to compare Gemini 2.5 Flash \texttt{Nothinking} versus \texttt{Thinking} and their successors, Gemini 3 Flash \texttt{Nothinking} versus \texttt{Thinking}. 
First, immature test-time reasoning induces performance degradation in code-oriented visual generation tasks. As reported in Tab. \ref{metric::image}, the integration of reasoning in Gemini 2.5 Flash-TK triggers a decline in both execution rates and structural similarity across all four programming languages.
We suspect that this effect originates from an over-thinking bias inherent in early models, which increases the risk of compilation collapse and causes the generated visualizations to deviate from the standard visual priors of the ground truth. 
Conversely, the Gemini 3 Flash series demonstrates a remarkable generational improvement in reasoning mechanisms.
Although activating reasoning capability decreases pixel-level image metrics and code similarity metrics, the human-aligned perceptual fidelity metric, MMOS, achieves large improvements in Python (from 76.69 to 78.51), Matlab (from 68.40 to 74.99), and R (from 69.76 to 73.01). 
This divergence reflects the nature of reasoning in visual code generation. Rather than relying on rote memorization of boilerplate syntax and default API aesthetics, reasoning models synthesize charts through complex, explicit intermediate logic. Although such an approach incurs mathematical penalties under strict string-matching and absolute pixel-level metrics, it ultimately produces visualizations of superior logical coherence, information completeness, and semantic correctness, thereby earning higher recognition within the perception-based MLLM evaluation framework.

{\bf Model Size Analysis.}
Fig.~\ref{modelsize} presents scatter plots of model size versus MMOS across four languages. We observe that increasing the parameter size generally yields consistent performance improvements within the same model family. 
In both the Qwen3-VL series (8B, 32B, 235B) and the InternVL3.5 series (8B, 38B), the logarithmic expansion of model size produces distinct upward trend lines in their MMOS across Python, Matlab, and R.
As one of the largest evaluated open-source models, Kimi-K2.5 consistently operates at or near the performance ceiling across all four languages, representing the upper bound achievable through pure parameter scaling.
Furthermore, we observe a counterintuitive inverse scaling phenomenon in the Qwen3.5 and GLM series, where generative performance degrades as model size increases. We hypothesize that this anomaly arises because their visual perception capabilities fail to scale proportionally with the expanding language backbone, creating a multimodal bottleneck in coordinate-based rendering tasks. Interestingly, a notable exception emerges in the LaTeX domain. Since Latex is a declarative and descriptive language, strengthening the language backbone can effectively compensate for lagging visual priors. In this scenario, advanced textual reasoning translates directly into improved geometric reproducibility, demonstrating that language scaling remains highly beneficial for purely logic-driven, text-intensive rendering paradigms.

\begin{figure*}[t!]
\centerline{\includegraphics[width=1\textwidth]{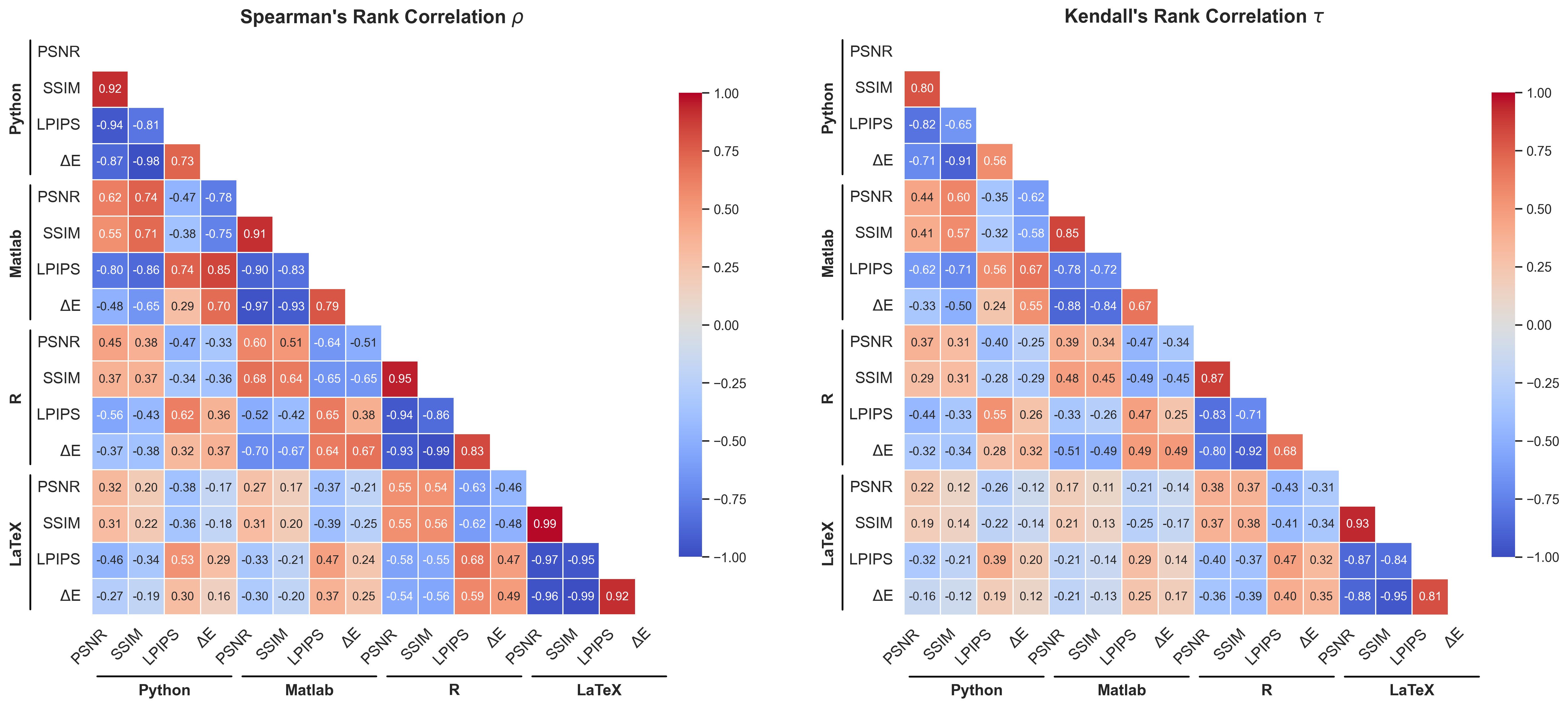}}
\caption{Correlation analysis of image-oriented metrics.}
\label{correlation::betweenimgmetric}
\end{figure*}

\begin{figure*}[t!]
\centerline{\includegraphics[width=1\textwidth]{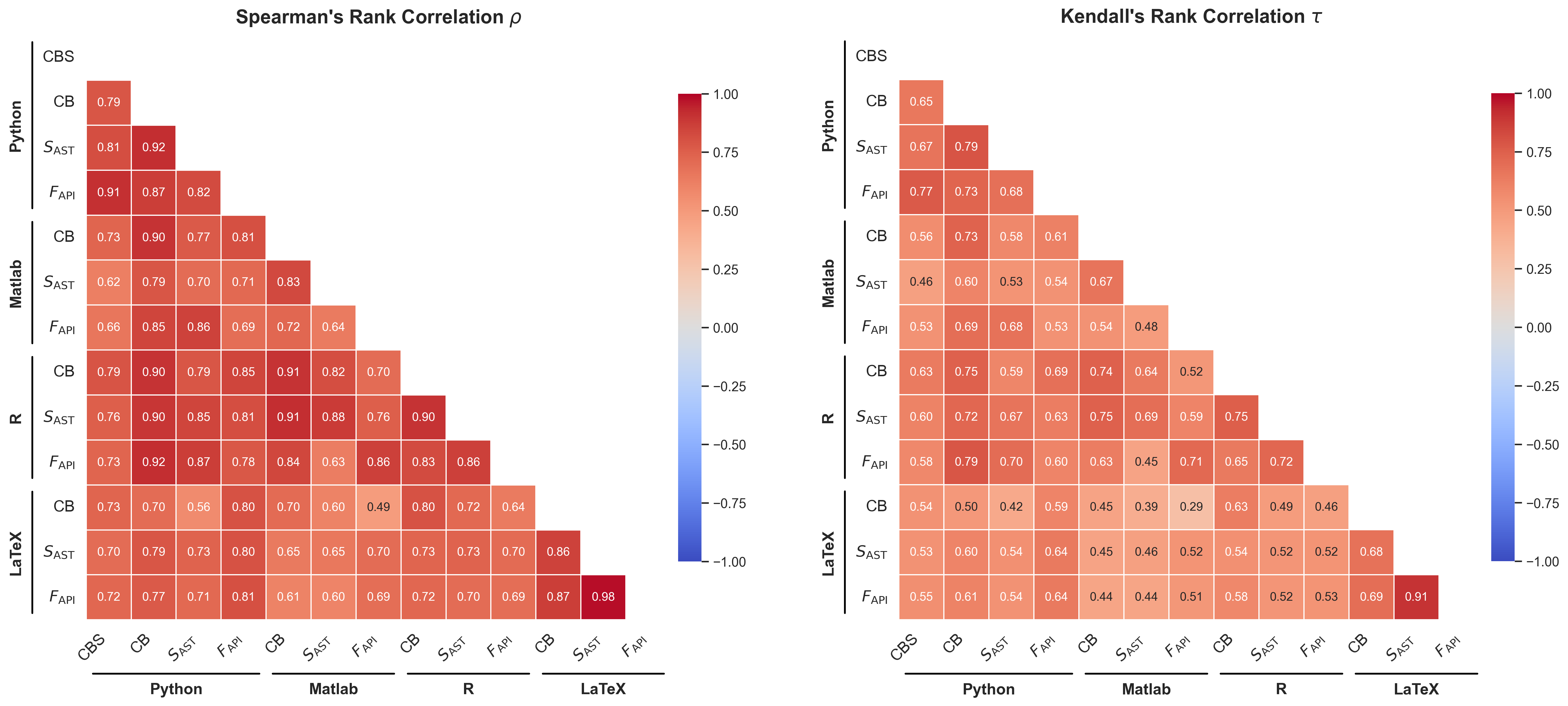}}
\caption{Correlation analysis of code-oriented metrics.}
\label{correlation::betweencodemetric}
\end{figure*}

\begin{table*}[!t]
  \caption{{\bf Correlation between image-oriented metrics and MMOS}. Given that code-oriented metrics primarily quantify similarity and thus do not directly reflect final graphical fidelity, we exclude them in this experiment.}
\renewcommand\arraystretch{1.15}
\belowrulesep=0pt\aboverulesep=0pt
  \label{correlation::imgMMOS}
  \centering
  \resizebox{1\linewidth}{!}{\begin{tabular}{l|cccc|cccc|cccc|cccc}
    \hline
    \multirow{2}{*}{\bf Model}     & \multicolumn{4}{c|}{\bf Python}     & \multicolumn{4}{c|}{\bf Matlab}&\multicolumn{4}{c|}{\bf R}&\multicolumn{4}{c}{\bf Latex} \\
    \cmidrule{2-4}\cmidrule{5-7}\cmidrule{8-10}\cmidrule{11-17}
&PSNR&SSIM&LPIPS&$\Delta E$&PSNR&SSIM&LPIPS&$\Delta \mathrm{E}$&PSNR&SSIM&LPIPS&$\Delta E$&PSNR&SSIM&LPIPS&$\Delta E$\\
    \hline
    Spearman's $\rho$&0.7687&0.7856&-0.7380&-0.7730&0.8035&0.7617&-0.8915&-0.5290&0.6783&0.7180&-0.7423&-0.7278&0.5313&0.5462&-0.6269&-0.5235   \\
    Kendall's $\tau$&0.5725&0.5621&-0.5699&-0.5435&0.6232&0.5725&-0.7441&-0.7104&0.4928&0.5554&-0.5554&-0.5870&0.3333&0.3376&-0.4335&-0.3406   \\
    \hline
  \end{tabular}}
\end{table*}

\subsection{Correlation Between Different Metrics}
To validate the rationality of our evaluation protocol, we conduct a comprehensive meta-evaluation of the intra-metric correlation, comparing MMOS against objective image-oriented metrics.
We use Spearman's rank correlation $\rho$ and Kendall's rank correlation $\tau$ as indicators.
First, as shown in Fig. \ref{correlation::betweenimgmetric}, in the diagonal blocks of the four coding environments, PSNR and SSIM show consistently strong positive correlations, indicating that they capture overlapping low-level reconstruction errors with equivalent results. Second, the deep perceptual metric LPIPS and the color metric $\Delta E$ demonstrate strong negative correlation trends with PSNR and SSIM. It is worth noting that this negative correlation is not an absolutely perfect linear divergence (e.g., the Kendall's $\tau$ between $\Delta E$ and PSNR in R is -0.34), showing that LPIPS and $\Delta E$ are capable of detecting deep semantic misalignments and subtle color gamut shifts that are easily overlooked by traditional pixel metrics.
Meanwhile, we notice that the pronounced decay in cross-language correlations highlights that the multimodal figure reproduction capability of MLLMs is highly dependent on language-specific syntactic priors, substantiating the necessity of constructing benchmarks that span multiple languages.

In Fig. \ref{correlation::betweencodemetric}, we observe a high consistency during cross-language comparisons, in contrast to the heterogeneity exhibited by image-oriented metrics. This reveals that the general programming proficiency of frontier MLLMs has strong cross-environment transferability.
Tab. \ref{correlation::imgMMOS} further reports the correlation between image-oriented metrics and MMOS. Consistent with the above analysis, deep semantic feature-based LPIPS aligns better with human preferences compared to the others. 
Meanwhile, all image-oriented metrics demonstrate moderate to strong correlations with MMOS across four programming languages, validating the rationality of employing image-level visual fidelity metrics for figure reproduction assessment.

\begin{figure*}[t!]
\centerline{\includegraphics[width=1\textwidth]{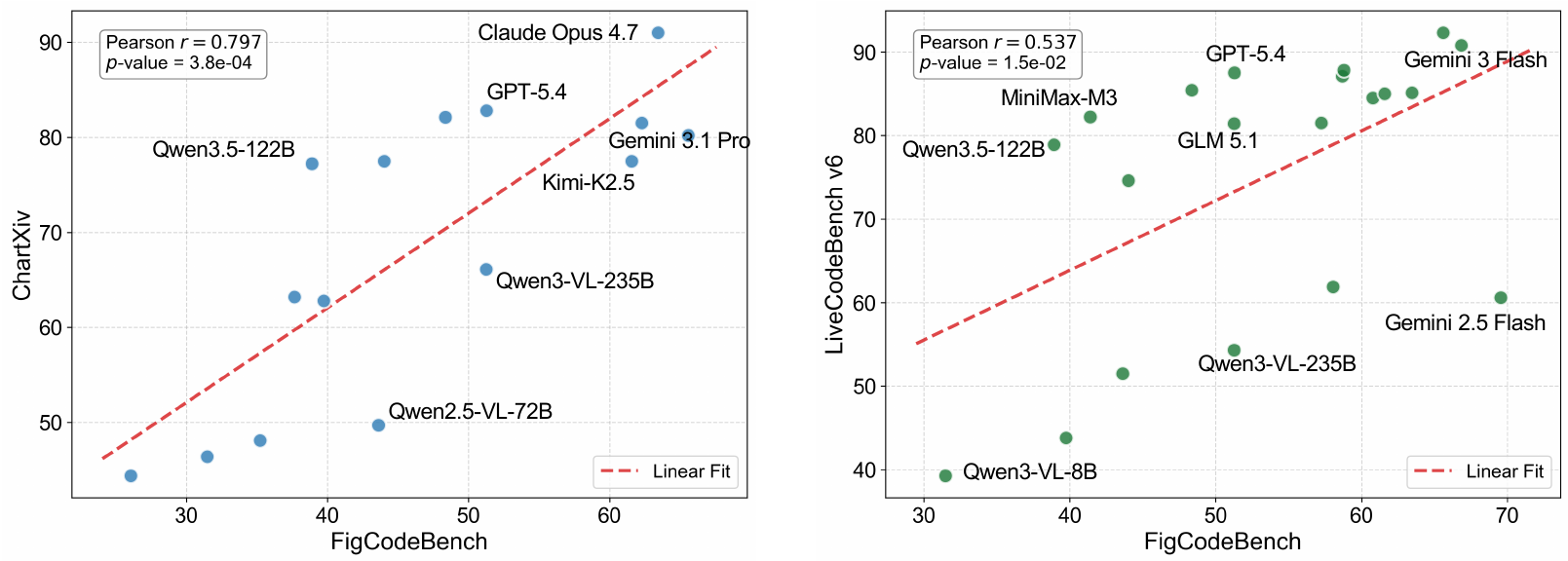}}
\caption{Performance correlation of FigCodeBench with benchmarks evaluating chart understanding and code generation capabilities.}
\label{correlation::other}
\end{figure*}

\subsection{Correlation Between FigCodeBench and Other Benchmarks}
To uncover the factors influencing FigCodeBench performance, we decompose the capabilities required for figure replication into chart understanding and code generation, and investigate the performance correlation between FigCodeBench and existing benchmarks targeting these two dimensions. Specifically, we adopt ChartXiv ({\it reasoning question}) \citep{wang2024charxiv} and LiveCodeBench ({\it v6}) \citep{zhenglivecodebench} as representative benchmarks for chart understanding and code generation, respectively. Pearson correlation coefficients $r$ are computed to quantify these relationships.
As depicted in \ref{correlation::other}, FigCodeBench shows a more stronger linear positive correlation with ChartXiv ($r=0.797$) than LiveCodeBench v6 ($r=0.537$), which indicates that proficiency in generic algorithmic programming does not directly translate to visual code generation. Unlike pure logic-driven coding tasks, reproducing figures requires deterministic coordinate spatial reasoning and precise geometric mapping. Therefore, the strong alignment with ChartXiv underscores that accurate visual comprehension serves as the indispensable basis for cross-modal image reproduction.
These results further demonstrate the necessity of FigCodeBench, which fills the gap in the evaluation of cross-modal perception and generation.

\begin{figure*}[t!]
\centerline{\includegraphics[width=1\textwidth]{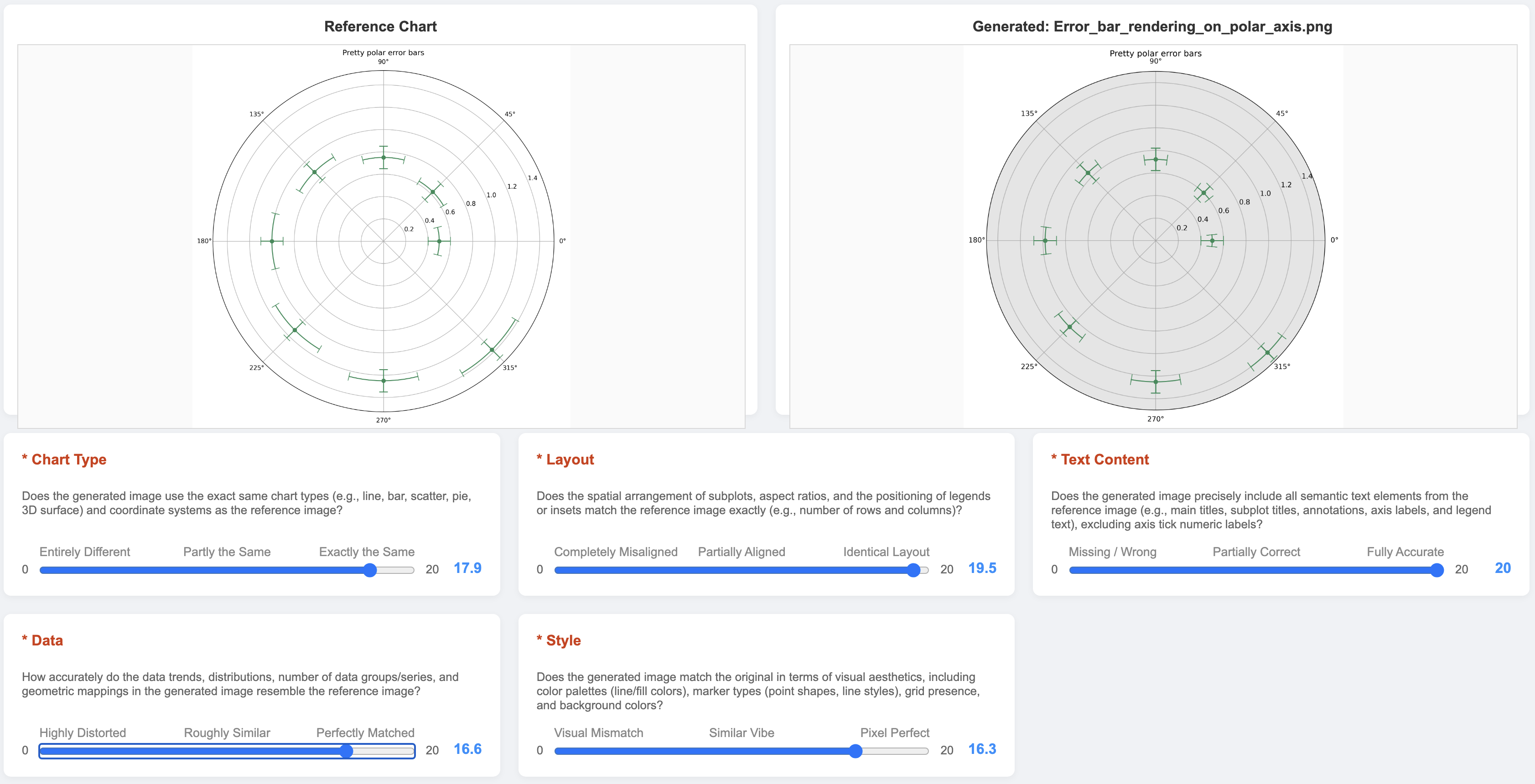}}
\caption{Screenshot of the user interface (UI) for subjective evaluation.}
\label{rating_UI}
\end{figure*}

\begin{table*}[!t]
  \caption{Pearson correlation coefficient $r$ between multi-level MMOS and human evaluation. $\mathrm{J_1}$, $\mathrm{J_2}$, and $\mathrm{J_3}$ represent GPT-5.1, GPT-5.6-Luna, and Gemini-3.5-Flash, respectively.}
\renewcommand\arraystretch{1.15}
\belowrulesep=0pt\aboverulesep=0pt
  \label{humaneval}
  \centering
  \resizebox{.7\linewidth}{!}{\begin{tabular}{c|cccccc}
    \hline
    {\bf Metric}& {\bf Chart Type} & {\bf Layout}&{\bf Text Content}&{\bf Data}&{\bf Style}&{\bf Overall} \\
    \hline
    $\mathrm{J_1}$&0.863&0.804&0.852&0.752&0.738&0.793\\
    $\mathrm{J_2}$&0.881&0.822&0.846&0.774&0.744&0.823\\
    $\mathrm{J_3}$&0.897&0.815&0.863&0.807&0.761&0.837\\
    \hline
    MMOS&0.883&0.817&0.855&0.783&0.747&0.826\\
    \hline
  \end{tabular}}
\end{table*}

\subsection{Human Evaluation}
\label{sec::humaneval}
To assess the reliability of the designed multi-level MLLM judges, we further conducted human evaluation for subjective consistency analysis.
Specifically, we recruited 12 experts from the university. Each of them has over three years of experience in using related programming languages and creating scientific diagrams. 
Following the same protocol as the MLLM judges, each human evaluator rates a generated-reference image pair on five dimensions (i.e., type, layout, style, text, and data), with scores ranging from 0 to 20. Subsequently, we calculate the Pearson correlation coefficient between the aggregate scores and the MMOS.
A total of 1,000 generated–reference image pairs are uniformly sampled from the four language-specific subsets of FigCodeBench.
Fig. \ref{rating_UI} provides a screenshot of the UI in the subjective evaluation. Before the annotation, we instructed all participants to have a clear and consistent understanding of all evaluated aspects and tested their eligibility via a 10-image pre-labeling.

As shown in Tab. \ref{humaneval}, this evaluation paradigm demonstrates strong alignment with human intuition across multiple dimensions. 
Notably, Gemini-3.5-Flash ($J_3$) achieves the highest correlation ($r=0.837$). By ensembling the preferences of different models, the aggregated MMOS achieves a strong overall correlation of 0.826, which effectively mitigates the evaluation biases introduced by a single weaker model (e.g., $J_1$ at 0.793), thereby providing a robust and highly human-aligned quantitative benchmark.
Moreover, we find that highly objective dimensions such as {\it chart type} and {\it text content} demonstrate a higher correlation with human judgments compared to the high-level {\it style} and rigorous {\it data} dimensions. This is partly attributed to the inherent variance of abstract attributes and partly attributed to the fundamental limitations of current visual encoders in performing fine-grained spatial mappings.
Besides, we assess the inter-rater reliability among human subjects using Krippendorff's alpha \citep{hayes2007answering}, yielding a value of $\alpha = 0.766$, which indicates strong agreement.

\begin{figure*}[t!]
\centerline{\includegraphics[width=1\textwidth]{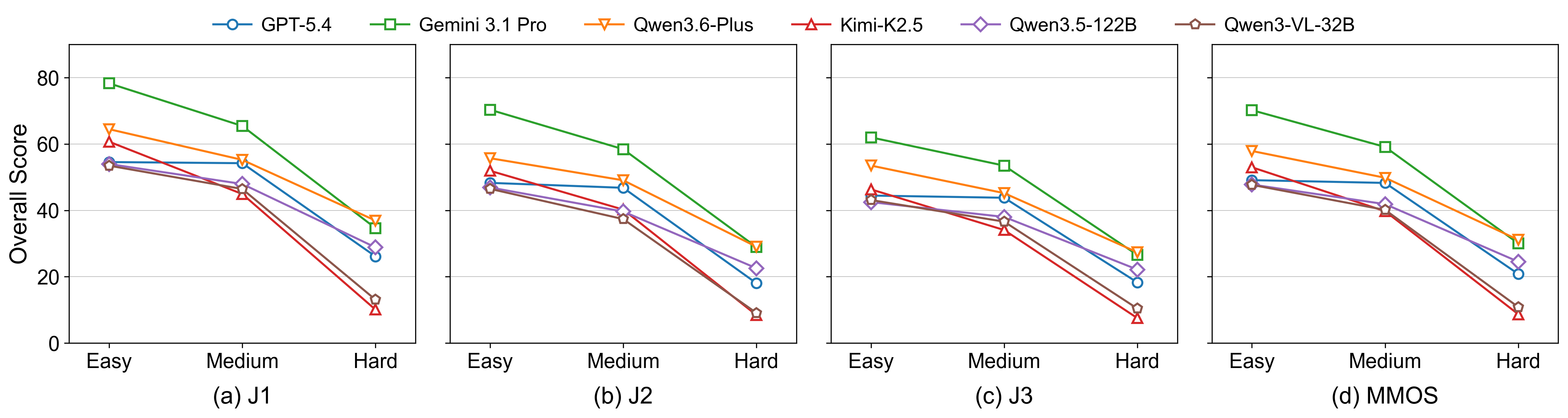}}
\caption{The MMOS of six representative models at different difficulty tiers.}
\label{difficultytier}
\end{figure*}

\subsection{Performance on Different Difficulty Tiers}
\label{exp::difficulty}
We provide the performance of six representative models, including GPT-5.4, Gemini 3.1 Pro, Qwen 3.6-Plus, Kimi-K2.5, Qwen3.5-122B, and Qwen3-VL-32B, across different difficulty tiers for FigCodeBench in Fig. \ref{difficultytier}.
We observe a consistent decline in performance across all languages as the difficulty increases. For example, the performance (MMOS) of Qwen3.6-Plus at easy, medium, and hard levels is 57.90, 49.82, and 31.11, respectively.
These results demonstrate that FigCodeBench is inherently challenging and confirm the efficacy of the established difficulty levels.
Additionally, we observe a more significant performance decrease in relatively weaker MLLMs, such as Qwen3.5-122B and Qwen3-VL-32B, than in top-tier models. Specifically, Qwen3-VL-32B suffers a relative decline of 77.2\%, from an average Easy score of 47.69 to a mere 10.85 on Hard tasks. In contrast, leading proprietary models show better resilience against figure complexity. For instance, Gemini 3.1 Pro and Qwen3.6-Plus maintain MMOS scores of 30.04 and 31.11 on Hard tasks, suffering relative drops of 57.21\% and 46.27\%, respectively, from the Easy level.
Meanwhile, we find that all models degrade only moderately from Easy to Medium, but much more sharply from Medium to Hard. This pattern suggests that the structural requirements in the Hard tier, such as extreme aspect ratios, intricate mathematical annotations, and densely overlapping geometric constraints, mark a level of complexity that current MLLMs still struggle with.

\begin{figure*}[t!]
\centerline{\includegraphics[width=.85\textwidth]{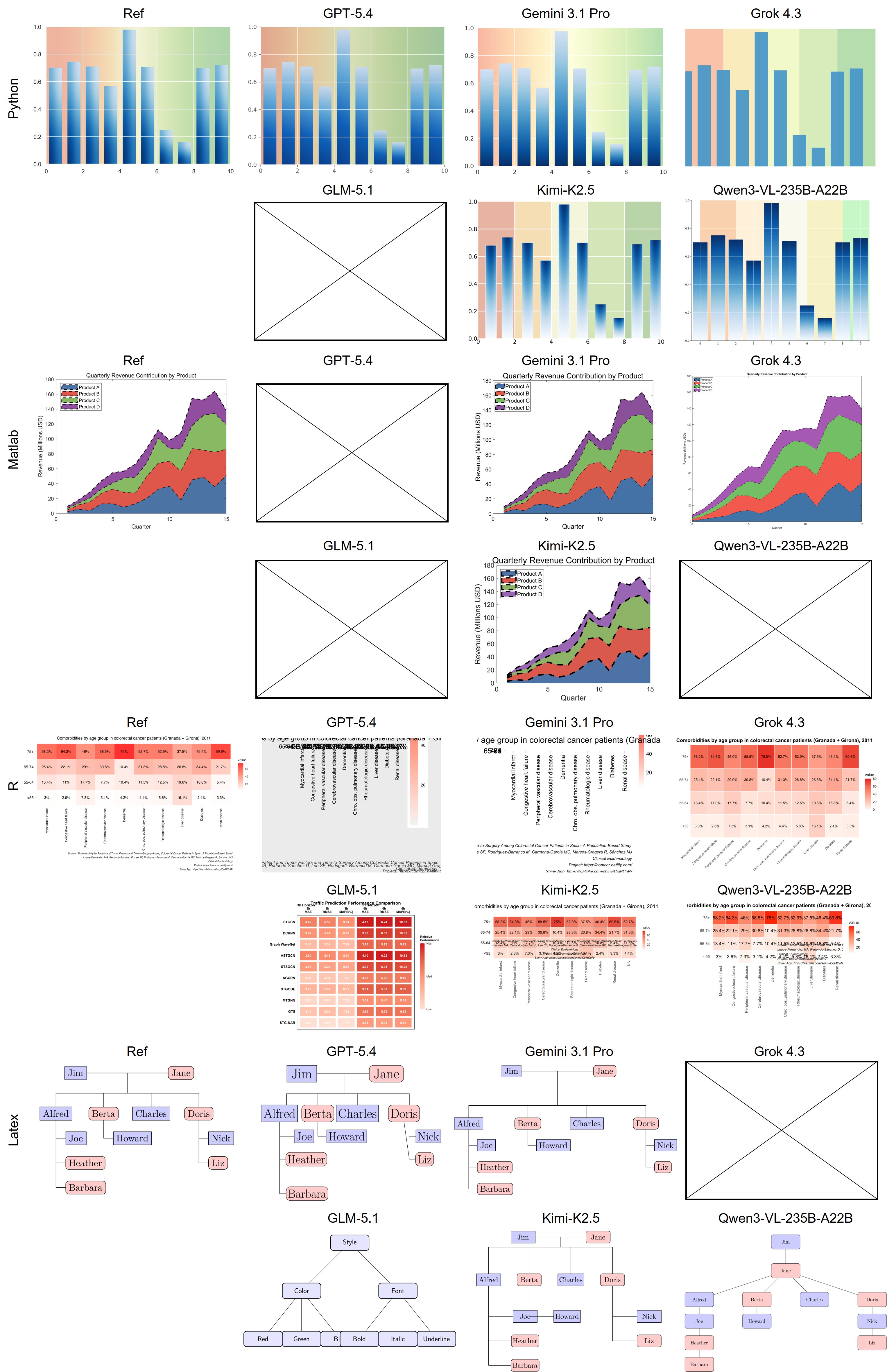}}
\caption{Comparison of figures reproduced by different models. We visualize results from six representative MLLMs, including GPT-5.4, Gemini 3.1 Pro, Grok 4.3, GLM-5.1, Kimi-K2.5, and Qwen3-VL-235B-A22B. The rectangle with diagonal lines indicates a failure in generation.}
\label{case-study}
\end{figure*}

\subsection{Case Study}
To further elucidate the limitations of MLLMs on FigCodeBench, we conduct an in-depth qualitative case study of the generation results from six representative models (GPT-5.4, Gemini 3.1 Pro, Grok 4.3, GLM-5.1, Kimi-K2.5, and Qwen3-VL-235B-A22B), as shown in Fig. \ref{case-study}. Through a cross-domain visual comparison against the ground-truth references, we identify several critical dimensions of deficiency in contemporary code-to-figure reproduction tasks as follows.
First, Gemini 3.1 Pro achieves the overall best visual reproduction performance in four programming languages. In the Python task, it perfectly reconstructs the complex continuous gradient background (transitioning from orange to green) and maintains accurate bar proportions. Similarly, in the Matlab area plot and Latex relationship tree, its alignment precision, typography, and structural topology are nearly indistinguishable from the reference images.
While several models successfully synthesize fundamental chart skeletons, they frequently struggle with micro-level visual styling. In the Python example, Kimi-K2.5 and Grok 4.3 fail to reproduce the smooth continuous gradient, erroneously discretizing the background into solid, segregated color blocks. Meanwhile, the color accuracy of Qwen3-VL-235B-A22B is even worse. In the Matlab stacked area chart, Grok 4.3 and Kimi-K2.5 generate prominent visual artifacts by excessively thickening or weakening the black dashed lines and altering the title proportions.
Second, current MLLMs exhibit severe degradation in spatial reasoning when managing dense text configurations or strict aspect ratios. In the R heatmap case, GPT-5.4, Kimi-K2.5, and Qwen3-VL-235B suffer a severe typographic failure, where a large amount of text elements overlap into an illegible black blob, entirely destroying chart readability.

As demonstrated by the previous quantitative experiments, challenging environments like Matlab and LaTeX induce frequent compilation failures, e.g., GLM-5.1 failing in both Python and MATLAB, and Grok 4.3 failing in Latex. In addition, in the Latex relationship tree, GLM-5.1 generates a topologically distinct and contextually irrelevant tree, replacing the reference human names with stylistic attributes such as "Style", "Color", and "Font". Furthermore, Qwen3-VL-235B incorrectly modifies the node fill styles and the global layout spacing of the tree.
These results suggest that, although current MLLMs have largely mastered basic data-to-geometry mapping, they still exhibit substantial structural weaknesses, such as spatial layout planning (e.g., collision avoidance and aspect-ratio control) and fine-grained aesthetic rendering (e.g., continuous colormaps).

\section{Conclusions}\label{sec::conclusion}
In this work, we introduce FigCodeBench, a rigorously curated benchmark designed to evaluate the unified perception and coding capabilities of MLLMs in figure reproduction. We construct FigCodeBench based on four primary principles, i.e., comprehensiveness, fidelity, reproducibility, and determinism.
FigCodeBench encompasses scientific drawing tasks with 6,194 figure-code pairs, spanning four programming languages and 7 function-focused figure categories that closely reflect common research scenarios. 
FigCodeBench is developed through a human-in-the-loop and semi-automatic process that consists of comprehensive data sources and multidimensional evaluation protocols, addressing the isolation characteristics in current image understanding and code generation benchmarks. 
Based on our framework, we conducted extensive experiments on 11 widely used proprietary and 13 open-source MLLMs, revealing their inferiority in precise figure reproduction and the inconsistency between figure reproduction and code generation capabilities.
We also investigate the performance correlation across metrics, programming languages, and ability dimensions, providing several valuable insights. 
We anticipate that FigCodeBench will drive future research in the field of perception-generation unified MLLMs.

{\bf Limitations.} 
Despite our best efforts, we acknowledge the following limitations in our study. First, since FigCodeBench is tailored to evaluate code generation for plotting, its findings may not fully generalize to other types of programming tasks.
Second, although our evaluation incorporates objective metrics with MLLM-as-a-judge semantics, it can inherently carry the risk of subtle alignment biases. For highly specialized domain figures, where a tiny geometric shift might fundamentally alter a physical or biological conclusion, general-purpose MLLM judges may lack the granular domain expertise required to penalize scientifically fatal, yet visually negligible, hallucinations.
Future work will explore the integration of domain-specific models to further calibrate the evaluation tier.

\section*{Declarations}
\begin{itemize}
\item Availability of data and materials: The FigCodeBench website will be available at \url{https://zijianchen98.github.io/FigCodeBench/}. The FigCodeBench codebase will be available at \url{https://github.com/zijianchen98/FigCodeBench}. The FigCodeBench dataset will be hosted and version-tracked on Hugging Face, and will be permanently accessible at \url{https://huggingface.co/datasets/CCZZJJ/FigCodeBench}.
We encourage the community to provide feedback and engage in scrutinizing the benchmark.
\item Competing interests: The authors declare no competing interests.
\item Funding: This work was supported in part by the National Key R\&D Program of China (2025ZD0124104) in collaboration with Shanghai Artificial Intelligence Laboratory, and in part by the Shanghai Municipal Special Program for Basic Research on General AI Foundation Models (Grant No. 2025SHZDZX025D09), in collaboration with Shanghai Artificial Intelligence Laboratory.
\end{itemize}

\newpage

\appendix
\section*{Appendix}

\section{Prompt Templates}
\label{app::templates}
In this section, we provide the prompt templates used for the experiment. 

\subsection{Prompt for Code Generation}
Note that we enforce uniform canvas geometries, e.g., DPI and aspect ratio configurations, to explicitly control the resolution of rendered figures.

\begin{tcolorbox}[boxrule=1.05pt]
\small
\texttt{\textbf{<|System Prompt|>}} \\
You are an expert developer who specializes in writing \textcolor{blue}{\texttt{\{code\_type\}}} code based on a given figure. \\ \\
\texttt{\textbf{<|User Prompt|>}} \\
I found a very nice figure in a paper, but there is no corresponding source code available. I need your help to generate the corresponding \textcolor{blue}{\texttt{\{code\_type\}}} code that can precisely reproduce the figure based on the image I provide.\\ \\
Requirement:\\
1. Ensure figsize=(width: \textcolor{blue}{\{width\}}, height: \textcolor{blue}{\{height\}}) to match the original image size.\\
2. Here is some auxiliary information to help you:\\
(a) The additional package dependencies include \textcolor{blue}{\{\texttt{package\_1}, ..., \texttt{package\_n}\}}. \\
(b) The dataset used in generating figure includes \textcolor{blue}{\{\texttt{dataset\_1}, ..., \texttt{dataset\_n}\}}. \\
(c) The random seed used in generating this figure is \textcolor{blue}{\texttt{\{random\_seed\}}}.\\
\\
Now, return only the code, without any additional explanations.\\
\end{tcolorbox}

\subsection{Prompt for Jury Models}
\begin{tcolorbox}[boxrule=1.05pt]
\small
\texttt{\textbf{<|System Prompt|>}} \\
You are an expert acting as a rigorous judge for evaluating visualization chart plots.
The first image (Reference Image) is the ground truth plot generated by various programming languages (e.g., Python, MATLAB, R, or LaTeX). The second image (AI-Generated Image) is rendered from the code generated by an AI assistant.
Your task is to evaluate how accurately the AI-generated plot replicates the reference plot. \\ \\
\texttt{\textbf{<|User Prompt|>}} \\
\# Scoring Methodology (Total: 100 points):
Evaluate the AI-generated image based on the following 5 criteria. Each criterion is worth up to 20 points.\\

1. Chart Types (20 points): Does the generated image use the exact same chart types (e.g., line, bar, scatter, pie, 3D surface) and coordinate systems as the reference image?\\
2. Layout (20 points): Does the spatial arrangement of subplots, aspect ratios, and the positioning of legends or insets match the reference image exactly (e.g., number of rows and columns)?\\
3. Text Content (20 points): Does the generated image precisely include all semantic text elements from the reference image (e.g., main titles, subplot titles, annotations, axis labels, and legend text), excluding axis tick numeric labels?\\
4. Data (20 points): How accurately do the data trends, distributions, number of data groups/series, and geometric mappings in the generated image resemble the reference image?\\
5. Style (20 points): Does the generated image match the original in terms of visual aesthetics, including color palettes (line/fill colors), marker types (point shapes, line styles), grid presence, and background colors?\\

\# Output Instructions:\\
Analyze the images step-by-step, then provide the final scores. You MUST output your response entirely in the following valid JSON format, without any surrounding markdown blocks or additional text: \\
\{ \\
  ``evaluation": \{\\
    ``chart\_types": \{``score": \textless int between 0 and 20\textgreater \},\\
    ``layout": \{``score": \textless between 0 and 20\textgreater\},\\
    ``text\_content": \{``score": \textless between 0 and 20\textgreater\},\\
    ``data": \{``score": \textless between 0 and 20\textgreater\},\\
    ``style": \{``score": \textless between 0 and 20\textgreater\}\\
  \},\\
  ``final\_total\_score": \textless sum of the 5 scores\textgreater \\
\}
\end{tcolorbox}

\end{document}